 \documentclass[pmlr,twocolumn,10pt]{jmlr} 

\usepackage{booktabs}
\usepackage{float}
\usepackage[load-configurations=version-1]{siunitx} 

\theorembodyfont{\upshape}
\theoremheaderfont{\scshape}
\theorempostheader{:}
\theoremsep{\newline}

\definecolor{PrincetonOrange}{HTML}{ff8f00}

\newcommand{\defeq}{\stackrel{\text{def}}{=}}

\newcommand{\D}{\mathcal{D}}

\def\x{\mathbf{x}}

\newcommand{\ignore}[1]{}
\newcommand{\email}[1]{\texttt{#1}}

\DeclareMathAlphabet{\mathbfsf}{\encodingdefault}{\sfdefault}{bx}{n}

\newcommand{\E}{\mathbb{E}}

\newcommand{\eps}{\varepsilon}

\let\oldtfrac\tfrac
\renewcommand{\tfrac}[2]{\smash{\oldtfrac{#1}{#2}}}

\let\nablaold\nabla
\renewcommand{\nabla}{\nablaold\mkern-2.5mu}

\title{Machine Learning for Mechanical Ventilation Control}

\author{
  \Name{Daniel Suo}\thanks{Google LLC}\thanks{Princeton University},
  \Name{Naman Agarwal}\footnotemark[1],
  \Name{Wenhan Xia}\footnotemark[1]\footnotemark[2],
  \Name{Xinyi Chen}\footnotemark[1]\footnotemark[2],
  \Name{Udaya Ghai}\footnotemark[1]\footnotemark[2],
  \Name{Alexander Yu}\footnotemark[1],
  \Name{Paula Gradu}\footnotemark[1],
  \Name{Karan Singh}\footnotemark[1]\footnotemark[2],
  \Name{Cyril Zhang}\footnotemark[1]\footnotemark[2],
  \Name{Edgar Minasyan}\footnotemark[1]\footnotemark[2], 
  \Name{Julienne LaChance}\footnotemark[2], 
  \Name{Tom Zajdel}\footnotemark[2], 
  \Name{Manuel Schottdorf}\footnotemark[2], 
  \Name{Daniel Cohen}\footnotemark[2], 
  \Name{Elad Hazan}\footnotemark[1]\footnotemark[2] 
}

\usepackage[compact]{titlesec}
\usepackage[belowskip=-10pt,aboveskip=0pt,font=footnotesize,labelfont=bf]{caption}

\usepackage{floatflt}

\begin{document}

\maketitle

\begin{abstract}
Mechanical ventilation is one of the most widely used therapies in the ICU. However, despite broad application from anaesthesia to COVID-related life support, many injurious challenges remain.

We frame these as a control problem: ventilators must let air in and out of the patient's lungs according to a prescribed trajectory of airway pressure. Industry-standard controllers, based on the PID method, are neither optimal nor robust.

Our data-driven approach learns to control an invasive ventilator by training on a simulator itself trained on data collected from the ventilator. This method outperforms popular reinforcement learning algorithms and even controls the physical ventilator more accurately and robustly than PID.

These results underscore how effective data-driven methodologies can be for invasive ventilation and suggest that more general forms of ventilation (e.g., non-invasive, adaptive) may also be amenable.

\end{abstract}

\section{Introduction}


Mechanical ventilation is a widely used treatment with applications spanning anaesthesia \citep{coppola2014protective}, neonatal intensive care \citep{van2019modes}, and life support during the current COVID-19 pandemic \citep{meng2020intubation, wunsch2020mechanical, mohlenkamp2020ventilation}. This life-sustaining treatment has two common modes: invasive ventilation, where the patient is fully sedated, and assist-control ventilation, where the patient can initiate breaths \citep{patientventasynch}. 

Even though mechanical ventilation has been deployed in ICUs for decades, several challenges remain that can lead to ventilator-induced lung injury (VILI) for patients \citep{vili}. In pressure-support ventilation, a form of assist-control ventilation, evidence suggests that a combination of high peak pressure and high tidal volume can lead to tissue injury in the lung \citep{vili_1}. Pressure-support ventilation also suffers from patient-ventilator asynchrony, where the patient's breathing pattern does not match the ventilator's, and can result in hypoxemia (low level of blood oxygen), cardiovascular compromise, and patient discomfort \citep{asynchrony}.

However, the risk of developing VILI depends not only on factors related to the ventilator, but also on intrinsic characteristics of the patient's lung \citep{vili}. These characteristics usually cannot be directly observed, so trained clinicians must continuously monitor the patient. Given the highly manual process of mechanical ventilation, it is desirable to have control methods that can better track prescribed pressure targets and are robust to variations of the patient's lung.

\begin{figure}[!h]
     \centering
     \includegraphics[width=0.45\textwidth]{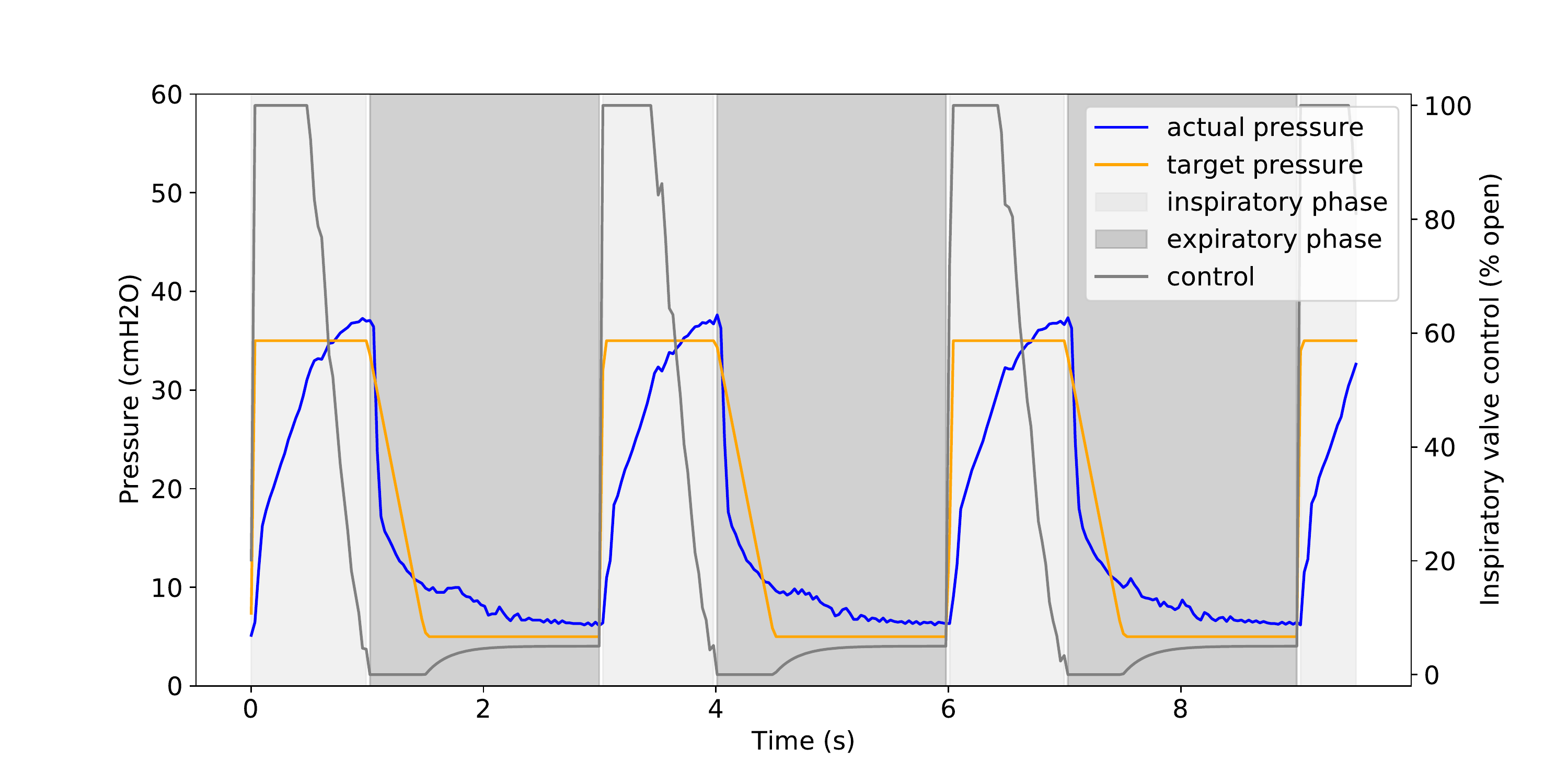}
     \caption{An example run of three breaths where PID (dark gray line) controls lung pressure (blue line) according to a prescribed target waveform (orange line).}
     \label{fig:tracking}
 \end{figure}

Motivated by this potential to improve patient health, we focus on pressure-controlled invasive ventilation (PCV) \citep{rittayamai2015pressure} as a starting point. In this setting, an algorithm controls two valves that let air in and out of a patient's lung according to a target waveform of lung pressure (see Figure \ref{fig:tracking}). We consider the control task only on ISO-standard \citep{ISO68844} artificial lungs.

\paragraph{State of the art.} 
Despite its importance, ventilator control has remained largely unchanged for years, relying on PID \citep{pid2} controllers and similar variants to track patient state according to a prescribed target waveform. 
However, this  approach is not optimal in terms of tracking---PID can overshoot, undershoot, and exhibit ringing behavior for certain lungs. It is also not sufficiently robust---ventilators are carefully tuned during design, manufacture, and maintenance \citep{ziegler1942optimum,chen2012control} and any changes in ventilator dynamics (e.g., tubing, response delay), environment (e.g., atmospheric pressure), or patient must be accounted for and continuously monitored by trained clinicians via various physical controls on the ventilator \citep{rees2006using}. 



\begin{figure}[!h]
     \centering
     \includegraphics[width=0.5\textwidth]{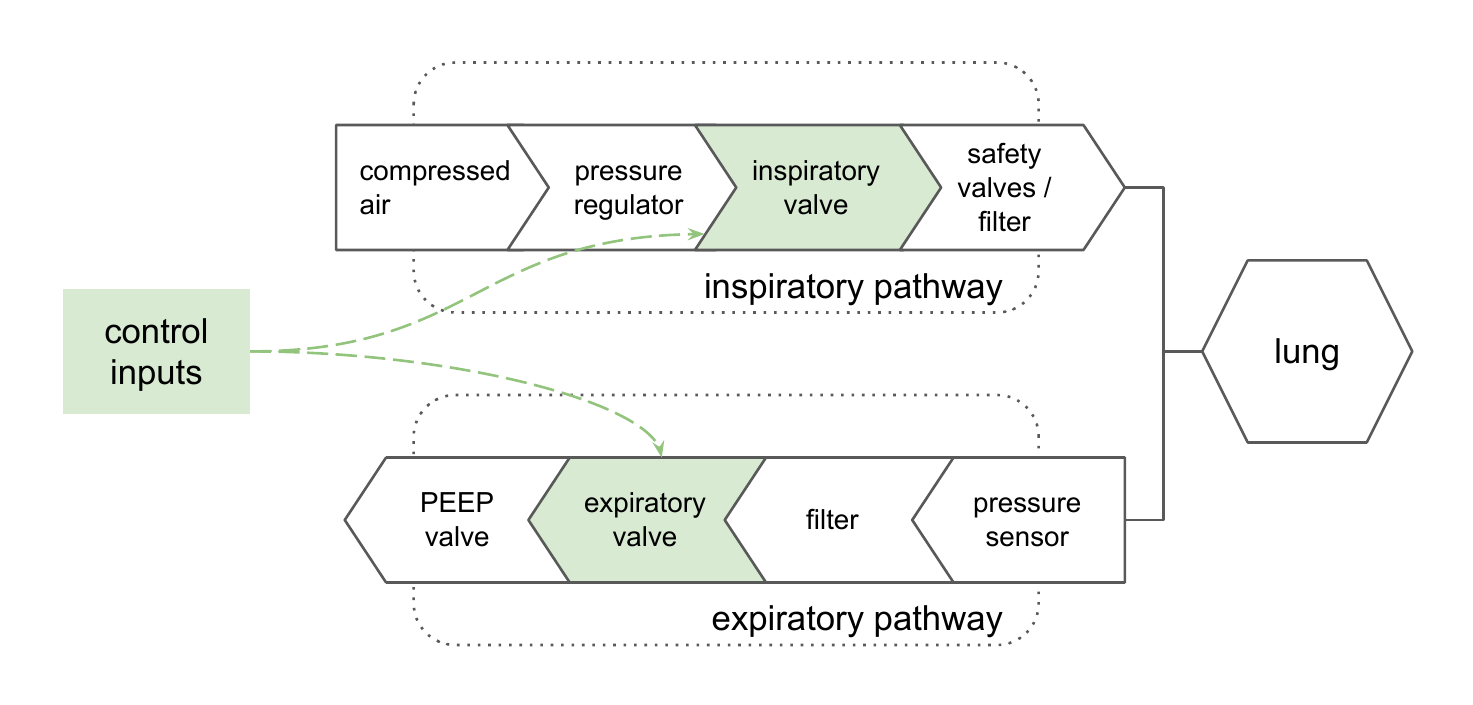}
     \caption{A simplified respiratory circuit showing the airflow through the inspiratory pathway, into and out of the lung, and out the expiratory pathway. We shade the components that our algorithms can control in green.}
     \label{fig:circuit}
 \end{figure}%

\paragraph{Challenges of ventilator control.}


A ventilator controller must adapt quickly and reliably across the spectrum of clinical conditions, which are only indirectly observable given a single measurement of pressure. 
A model that is highly expressive may learn the dynamics of the underlying systems more precisely and thus adapt faster to the patient's condition. However, such models usually require a large amount of data to train, which can take prohibitively long to collect by purely running the ventilator. We opt instead to learn a simulator to generate artificial data, though learning such a simulator for a partially observed non-linear system is itself a difficult problem.






\subsection{Our contributions} 

We present better-performing, more robust results and present resources for future researchers. Specifically, 
\begin{enumerate}

    \item We demonstrate that learning a controller as a neural network correction to PID outperforms its uncorrected counterpart (optimality).
    
    \item We show that a single learned controller trained on several ISO lung settings outperforms the PID controller that performs best across the same settings (robustness).
    
    \item We provide self-contained differentiable simulators for the ventilation problem. These simulators  reduce the entrance cost for future researchers to contribute to invasive mechanical ventilation. 
    
    \item We conduct a methodological study of reinforcement learning techniques, both model-based and model-free, including policy gradient, Q-learning and other variants. We conclude that model-based approaches are more sample and computationally efficient. 
    

\end{enumerate}


Of note, we limit our investigation to open-source ventilators. Control methods used by proprietary ventilators cannot be modified or assessed independently from their hardware and such equipment are cost-prohibitive for academic research.

We see this study as a preliminary investigation of machine learning for ventilator control. In future work, we hope to extend this methodology to non-invasive ventilation, pressure-support ventilation, and conduct clinical trials. 


\subsection{Related work}\label{sec:related}

The modern positive-pressure ICU mechanical ventilator dates back to the 1940s \citep{kacmarek11} with many open-source ventilator designs \citep{ventlist} published during the COVID-19 pandemic.

Yet at their core, ventilators all rely on controlling air in and out of an elastic lung via a respiratory circuit, as described in many physics-based models \citep{PMID:8420408}. Such simple operation masks the complexity of treatment \citep{chatburn2011closed} and recent work on augmenting PID controllers with adaptive methods \citep{9122946,PPR:PPR169448} have sought to address more advanced clinical needs. To the best of our knowledge, our data-driven approach of learning both simulator and controller is novel in this field.

\paragraph{Control and RL in virtual and physical systems.} Much progress has been made on learning dynamics when the dynamics themselves exist \emph{in silico}: MuJoCo physics \citep{hafner2019learning}, Atari games \citep{kaiser2019model}, and board games \citep{schrittwieser2020mastering}. Combining such data-driven models with either pixel-space or latent-space planning has been shown to be effective  for policy learning. \citep{ha2018recurrent} is an example of this research program for the deep learning era. Progress on deploying end-to-end learned agents (i.e. controllers) in the physical world is more limited in comparison, due to difficulties in scaling parallel data collection and higher variability in real-world data. \citep{bellemare2020autonomous} present a case study on autonomous balloon navigation using a Q-learning approach, rather than a  model-based one like ours. \citep{akkaya2019solving} use domain randomization with non-differentiable simulators for a difficult dexterous manipulation task.



\paragraph{System identification and residual policy learning. } System identification has been studied for decades in control and reinforcement learning,  
see e.g. \citep{schoukens2019nonlinear,billings1980identification} for nonlinear system identification.   Deep neural networks have been used to represent nonlinear dynamics, see e.g. \cite{helicopter}. Residual policy learning \citep{rpl} is a model-free analogue of our controller design: it learns a correction term on an initial, imperfect policy, and is shown to be more data-efficient than learning from scratch, especially for complex robotic tasks. More recently, concurrent work by \citep{pidcar} uses residual policy learning to improve PID for the car suspension control problem.

\paragraph{Multi-task reinforcement learning.} Part of our methodology has close parallels in multi-task reinforcement learning \citep{taylor2009transfer}, where the objective is to learn a policy that performs well across diverse environments. To make our controllers more robust, we optimize our policy simultaneously on an ensemble of learned models corresponding to different physical settings, similar to the work of \citep{rajeswaran2016epopt, chebotar2019closing} on robotic manipulation.

\paragraph{Machine learning for health applications.} Healthcare offers a multitude of opportunities and challenges for machine learning; for a survey, see \citep{ghassemi2020review}. Specifically, reinforcement learning and control have found numerous applications \citep{yu2020reinforcement}, and recently for weaning patients off mechanical ventilators \citep{prasad2017reinforcement,yu2019inverse,yu2020supervised}. As far as we know, there is no prior work on improving the control of ventilators using machine learning.

\section{Scientific background}
\label{sec:background}

\subsection{Control of dynamical systems}

We begin with some formalisms of the control problem. A partially-observable discrete-time dynamical system is given by the following equation: 
$$ x_{t+1} = f(x_t, u_t), o_{t+1} = g(x_{t+1})$$
where  $x_t$ is the underlying state of the dynamical system, $o_t$ is the observation of the state available to the controller, $u_t$ is the control input and $f,g$ are the transition function and observation functions respectively. Given a dynamical system, the control problem is to minimize the sum of cost functions over a long-term horizon:
\begin{align*}
& \min_{u_{1:T} }  \sum_{t=1}^T c_t(x_t, u_t) \quad \text{s.t.}\;\; x_{t+1} = f_t(x_t, u_t).
\end{align*}

This problem is in general computationally intractable, and theoretical guarantees are available for special cases of dynamics (notably linear dynamics) and perturbations. For an in-depth exposition on the subject, see the textbooks by \citet{Bertsekas17,kemin,tedrake}. 

\paragraph{PID control.} A ubiquitous technique for the control of dynamical systems is the use of linear error-feedback controllers, i.e. policies that choose a control based on a linear function of the current and past errors vs. a target state. That is,
$$ u_{t+1} = \sum_{i=0}^k \alpha_i \eps_{t-i} , $$
where $\eps_t = x_t - {x}^{\star_t}$ (or $\eps_t = o_t - {o}^{\star_t}$ if the system is partially-observable) is the deviation from the target state at time $t$, and $k$ represents the history length of the controller. PID applies a linear control with \emph{proportional}, \emph{integral}, and \emph{differential} coefficients,
$$ u_t = \alpha \eps_{t} + \beta \sum_{i=0}^k \eps_{t-i} + \gamma (\eps_{t} - \eps_{t-1}) .  $$
This special class of linear error-feedback controllers, motivated by physical laws, is a simple, efficient and widely used technique  \citep{PID1}. It is currently the industry standard for (open-source) ventilator control.  

\subsection{The physics of ventilation}
\label{subsec:physics}
In invasive ventilation, the ventilator is connected to a patient's main airway, and applies pressure in a cyclic manner to simulate healthy breathing. During the inspiratory phase, the target applied pressure increases to the peak inspiratory pressure (PIP). During the expiratory phase, the target decreases to the positive-end expiratory pressure (PEEP), maintained in order to prevent the lungs from collapsing. The PIP and PEEP values, along with the durations of these phases, define the time-varying target \emph{waveform}, specified by the clinician.


The goal of ventilator control is to regulate the pressure sensor measurements to follow the target waveform $p_t^{\star}$ via controlling the air-flow into the system which forms the control input $u_t$. As a dynamical system, we can denote the underlying state of the ventilator-patient system as $x_t$ evolving as $x_{t+1} = f(x_t, u_t),$ for an unknown $f$ and the pressure sensor measurement $p_t$ is the observation available to us. The cost function can be defined to be a measure of the deviation from the target; e.g. the absolute deviation $c_t(p_t, u_t) = |p_t - p_t^{\star}|$. The objective is to design a controller that minimizes the total cost over $T$ time steps. 

A ventilator needs to take into account the structure of the lung to determine the optimal pressure to induce. Such structural factors include \textit{compliance} ($C$), or the change in lung volume per unit pressure, and \textit{resistance} ($R$), or the change in pressure per unit flow.

\paragraph{Physics-based model.} A simplistic formalization of the ventilator-lung dynamical system can be derived from the physics of a connected two-balloon system, with a \emph{latent state} $v_t$ representing the volume of air inside the lung. The dynamics equations can be written as
$$v_{t+1} = v_t + u_t \cdot \Delta_t$$ $$p_t = p_0 + \left( 1 - \left( \frac{r_t}{r_0} \right)^6 \right) \cdot \frac{1}{r_t r_0^2}, \;\;r_t = \left( \frac{3 v_t}{4 \pi} \right)^{1/3},$$ where $p_t$ is the measured pressure, $v_t$ is volume, $r_t$ is radius of the lung, $u_t$ is the input air flow rate, and $\Delta_t$ is the time increment. $u_t$ originates from a pressure difference between lung-pressure $p_t$ and supply-pressure $p_\text{supply}$, regulated by a valve: $u_t = \frac{p_\text{supply} - p_t}{R_\text{in}}$. The resistance of the valve is $R_\text{in}\propto 1/d^4$ (Poiseuille's law) where $d$, the opening of the valve, is controlled by a motor. The constants $p_0,r_0$ depend on both the lung and ventilator.  In \cite{nadeem_2021}, several physics-based models are benchmarked, showing errors that are an order of magnitude larger than what can be achieved with a data driven approach.  While the interpretability of such models is appealing, their low fidelity is prohibitive for offline reinforcement learning.

\subsection{Challenges and benefits of a model-based approach}

The physics-based dynamics models described above are highly idealized, and are suitable only to provide coarse predictions for the behaviors of very simple controllers. We list some sources of error arising from using physics equations for model-based control:
\begin{itemize}
    \item \emph{Idealization of physics:} oversimplifying fluid flow and turbulence via ideal incompressible gas assumptions; linearizing the dynamics of the lung and ventilator components.
    \item \emph{Lagged and partial observations:} assuming instantaneous changes to volume and pressure across the system. In reality, there are non-negligible propagation times for pressure impulses, delayed pressure feedback arising from lung elasticity, and computational latency.
    \item \emph{Underspecification of variability:} different patients' clinical scenarios, captured by the latent constants $p_0, r_0$, may intrinsically vary in more complex (i.e. higher-dimensional) ways.
\end{itemize}

Due to the reasons listed above, it is highly desirable to adopt a learned model-based approach in this setting because of its sample-efficiency and reusability. A reliable simulator enables much cheaper and faster data collection for training a controller, and allows us to incorporate multitask objectives and domain randomization (e.g. different waveforms, or even different patients). An additional goal is to make the simulator \emph{differentiable}, enabling direct gradient-based policy optimization through the system's dynamics (rather than stochastic estimates thereof).

We show that in this partially-observed (but single-input single-output) system, we can query a reasonable amount of training data in real time from the test lung, and use it offline to learn a differentiable simulator of its dynamics (\emph{``real2sim''}). Then, we complete the pipeline by leveraging interactive access to this simulator to train a controller (\emph{``sim2real''}). We demonstrate that this pipeline is sufficiently robust that the learned controllers can outperform PID controllers tuned directly on the test lung.
\section{Experimental Setup}
\label{sec:hardware}

To develop simulators and control algorithms, we run mechanical ventilation tasks on a physical test lung \citep{ingmar_medical_2020} using the open-source ventilator designed by Princeton University's People's Ventilator Project (PVP) \citep{pvp2020}.

\subsection{Physical test lung}
\begin{figure}[!h]
    \centering
    \includegraphics[width=7cm]{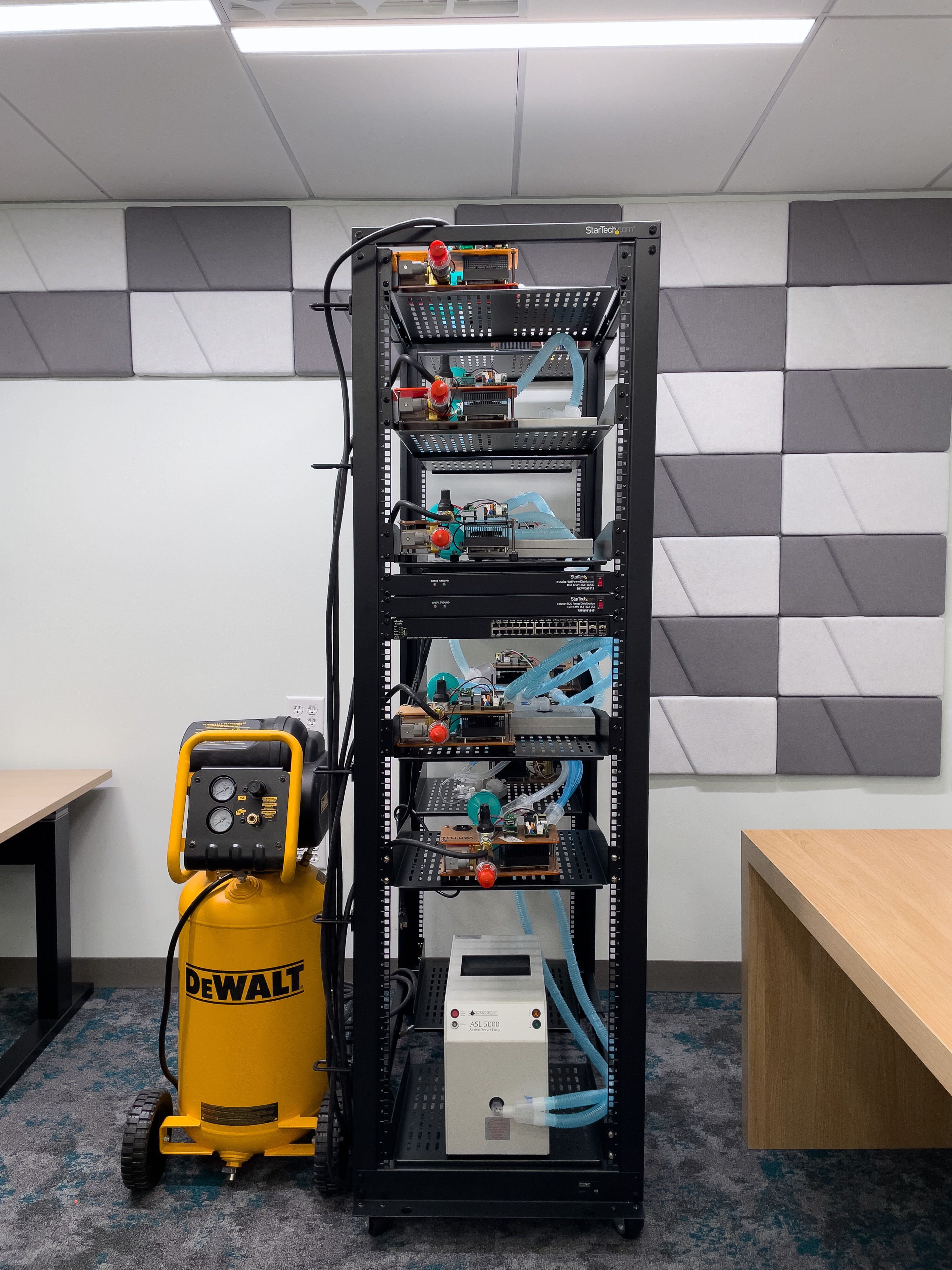}
    \caption{The ventilator cluster we constructed to run our experiments, featuring 10 ventilators, 4 air compressors, and 2 control servers. Each ventilator is re-calibrated after each experimental run for consistency across ventilators and over time.}

    \label{fig:vent-farm}
\end{figure}

For our experiments, we use the commercially-available adult test lung, ``QuickLung'', manufactured by IngMar Medical. The lung has three lung compliance settings ($C=\{10, 20, 50\}$ mL/cmH2O) and three airway resistance settings ($R=\{5, 20, 50\}$ cmH2O/L/s) for a total of 9 settings, which are specified by the ISO standard for ventilatory support equipment \citep{ISO68844}. An operator can change the lung's compliance and resistance settings manually. We connect the test lung to the ventilator via a respiratory circuit \citep{canadian1986canadian, parmley1972disposable} as shown in Figure \ref{fig:circuit}. Figure \ref{fig:vent-farm} shows a snapshot of our hardware setup.


\subsection{Mechanical ventilator}
There are many forms of ventilator treatment. In addition to various pressure target trajectories, clinicians may want to focus on other factors, such as volume and flow \citep{chatburn2007classification}. The PVP ventilator focuses on targeting pressure for a completely sedated patient (i.e., the patient does not initiate any breaths) and comprises two pathways (see Figure \ref{fig:circuit}): (1) the inspiratory pathway that regulates airflow into the lung and (2) the expiratory pathway for airflow out of the lung. A software controller is able to adjust one valve for each pathway. The inspiratory valve is a proportional control flow valve that allows control in a continuous range from fully closed to fully open. The expiratory valve is a binary on-off valve that only permits zero or maximum airflow.

To prevent damage to the ventilator and/or injury to the operator, we implement software overrides that abort a given run: 1) if pressure or volume in the lung exceeds certain thresholds, 2) if tubing disconnects, or 3) if there is significant software delay. The PVP pneumatic design also includes a safety valve in case software overrides fail.




\subsection{Abstraction of the simulation task}
\begin{figure}
    \centering
    \includegraphics[width=8cm]{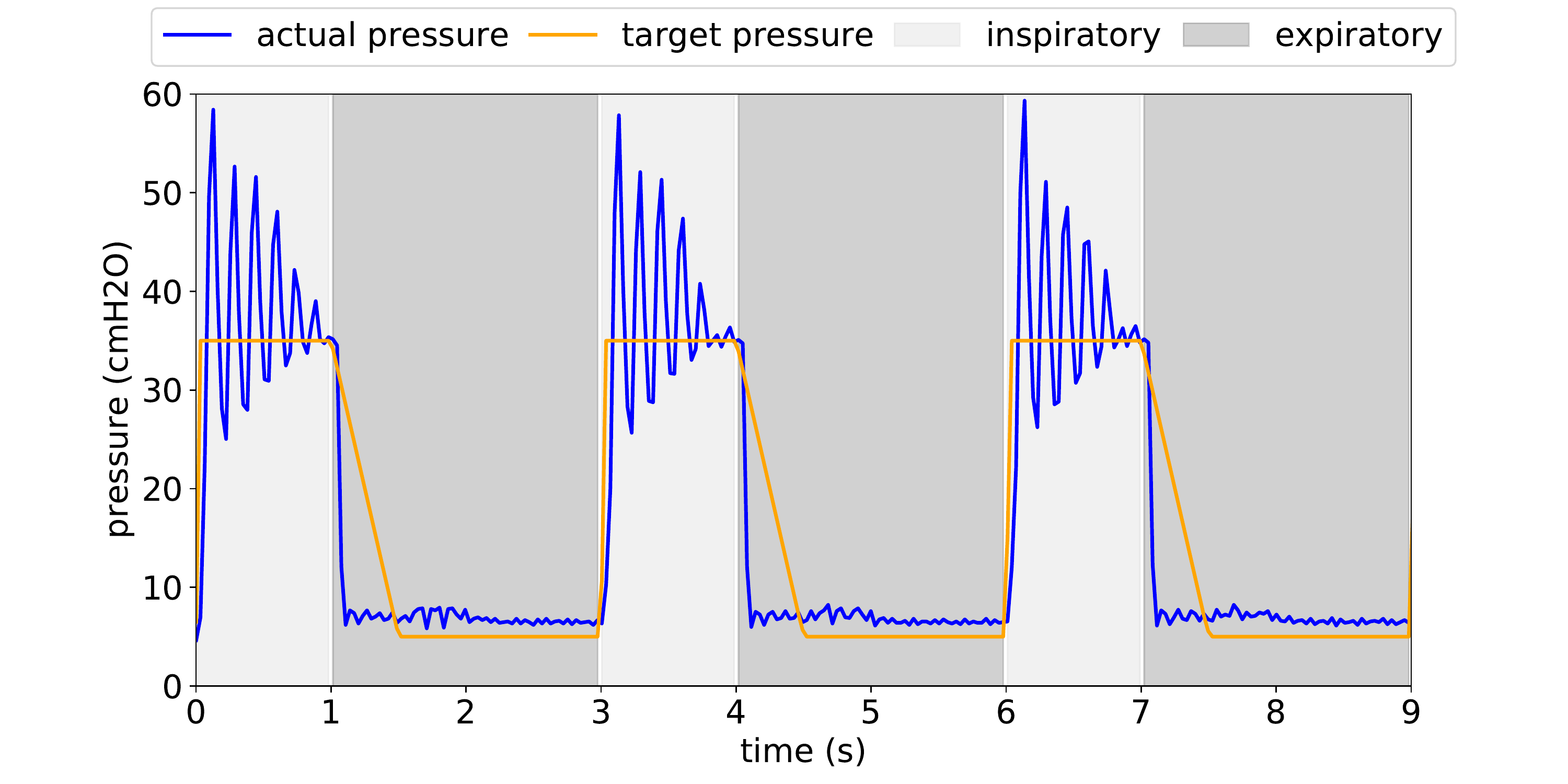}
    \caption{PID controllers exhibit their suboptimal behavior (under- or over-shooting, ringing) primarily during the inspiratory phase. Note that we use a hard-coded controller during expiratory phases to ensure safety. This choice does not affect our results.}
    \label{fig:lung-task}
\end{figure}

We treat the mechanical ventilation task as episodic by separating each inspiratory phase (e.g., light gray regions in Figure \ref{fig:lung-task}) from the breath timeseries and treating those as individual episodes. This approach reflects both physical and medical realities. Mechanically ventilated breaths are by their nature highly regular and feature long expiratory phases (dark gray regions in Figure \ref{fig:lung-task}) that end with the ventilator-lung system close to its initial state, thereby justifying the episodic nature. Further, the inspiratory phase is indeed the most relevant to clinical treatment and the harder regime to control with prevalent problems of under- or over-shooting the target pressure and ringing. Naturally thus, we attempt to learn a simulator for the ventilator-lung dynamics for the inspiratory phase. To this end repeated episodes of inspiratory phases are thus simplified, faithful units of training data.

\section{Learning a data-driven simulator} \label{sec:sim}

With the hardware setup outlined in Section~\ref{sec:hardware}, we have a physical system suitable for benchmarking, in place of a true patient's lung. In this section, we present our approach to learning a simulator for the inspiratory phase of this ventilator-lung system, subject to the practical constraints of real-time data collection. Two main considerations drive our simulator training and evaluation design:  

First, the evaluation for any simulator can only be performed using a {\bf black-box metric}, since we do not have explicit access to the system dynamics, and existing physics models are poor approximations to the empirical behavior. 
    
Second, the dynamical system we simulate is very challenging for a comprehensive simulation covering all modalities and in particular exhibits chaotic behavior in boundary cases. Therefore, since the end goal for the simulator is better control, we only evaluate the simulator for ``reasonable" scenarios that are relevant to the control task.

\subsection{Black-box simulator evaluation}\label{sec:sim_metric}

    
The class of learned simulators we consider are deep neural networks and thus in addition to the lack of explicit access to system dynamics, the simulator dynamics are also complex non-linear operations. Thus we deviate from standard distance metrics (between the simulator and the true system) considered in the literature, such as \citet{ferns2005metrics}, as they explicitly involve the value function over states, transition probabilities or other unknown quantities. Rather, we consider metrics that are based on the evolution of the dynamics, as studied in \citet{vishwanathan2007binet}. 

However, unlike the latter work, we take into account the particular distribution over control sequences that we expect to search around during the controller training phase. We thus define the following distance between dynamical systems.
Let $f_1,f_2$ be two dynamical systems over the same state-action spaces. 
Let $\D$ be a distribution over sequences of controls denoted $\mathbf{u} = \{u_1,u_2,...,u_T\}$.
We define the {\bf open-loop distance} w.r.t. horizon $T$ and control sequence distribution $\D$ as
    \begin{align*}
    \|f_1 - f_2\|_{ol} \defeq \E_{\mathbf{u} \sim \D} \left[ \sum_{t=1}^T \| f_1(x_{t,1},u_t) - f_2(x_{t,2},u_t) \|  \right] . \end{align*}

    

We use the Euclidean norm over the states in the inner loop, although this can be generalized to any metric. Compared to metrics involving feedback from the simulator, the open-loop distance is a more reliable description of transfer, since it minimizes hard-to-analyze interactions between the policy and the simulator.

\begin{figure}
\centering
\includegraphics[width=8cm]{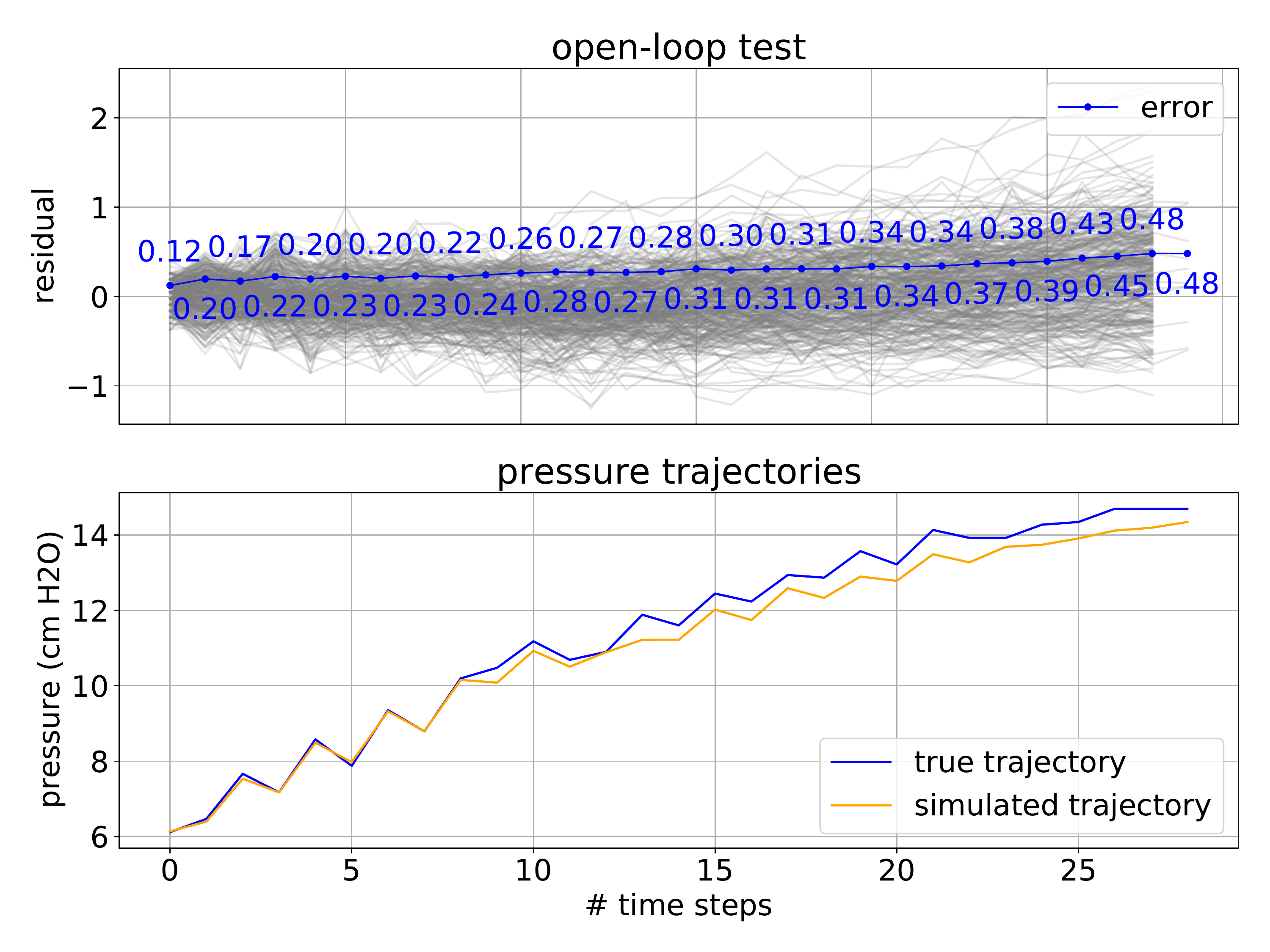}
\caption{Performance for a learned simulator for ventilator + a particular lung setting $R=5, C=50$. The upper plot shows the \textit{open-loop distance} and the lower shows a sample trajectory with a fixed sequence of controls (open-loop controls), both for lung setting . In the former, we see low error as we increase the number of steps we project and in the latter, we see that our simulated trajectory tracks the true trajectory quite closely.}\label{fig:sim_testing} 
\end{figure}

We evaluate our data-driven simulator using the open loop distance metric, and we illustrate a result in the top half of Figure~\ref{fig:sim_testing}. In the bottom half, we show a sample trajectory of our simulator and the ground truth. See Section \ref{sec:sim_model} for experimental details. 

\subsection{Data-collection via targeted exploration}\label{sec:safe_exp}
Motivated by the black-box metric described above, we focus on collecting trajectories comprising of control sequences and the measured pressure sequences upon the execution of the control sequences to form a training dataset. Due to safety and complexity issues, we cannot hope to exhaustively explore the space of all trajectories. Instead keeping the eventual control task in mind, we choose to explore trajectories \textit{near} the control sequence generated by a baseline PI controller. The goal is to have the simulator faithfully capture the true dynamics in a reasonably large vicinity of the optimal control trajectory on the true system. To this end, for each of the lung settings, we collect data by choosing a safe PID controller baseline and introducing random exploratory perturbations according to the following two policies:
\begin{enumerate}
    \item Boundary exploration: To the very beginning of the inhalation, add an additional control sampled uniformly from $(c^a_{\min}, c^a_{\max})$ and decrease this additive control linearly to zero over a time frame sampled randomly from $(t^a_{\min}, t^a_{\max})$; 
    \item Triangular exploration: sample a maximal additional control from a range $(c^b_{\min}, c^b_{\max})$ and an interval $(t^b_{\min}, t^b_{\max})$, within the inhalation. Start from $0$ additional control at time $t^b_{\min}$, increase the additional control linearly until $(t^b_{\min} + t^b_{\max}))/2$, and then decrease to $0$ linearly until $t^b_{\min}$.
\end{enumerate}

For each breath during data collection, we choose policy $(a)$ with probability $p_a$ and policy $(b)$ with probability $(1-p_a)$. The ranges in $(a)$ and $(b)$ are lung-specific. We give the exact values used in the Appendix.

\begin{figure}\centering
\includegraphics[width=8cm]{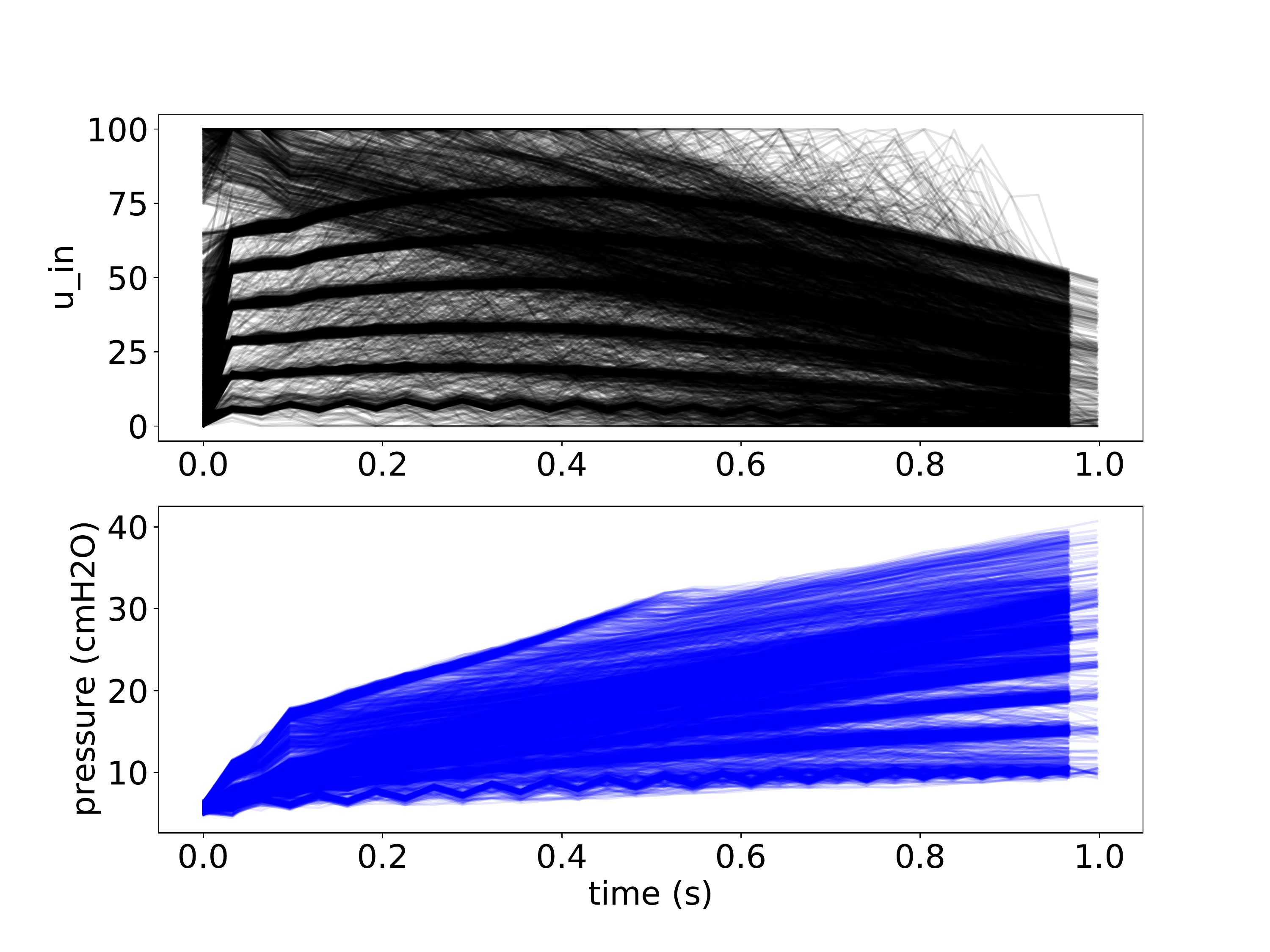}
\caption{We overlay the controls and pressures from all inspiratory phases in the upper and lower plots, respectively. From this example of the simulator training data (lung setting $R=5, C=50$), we see that we explore a wide range of control inputs (upper plot), but a more limited ``safe'' range around the resulting pressures.}\label{fig:exp_data}
\end{figure}

This protocol balances the need to explore a significant part of the state space with the need to ensure safety. The boundary exploration capitalizes on the fact that at the beginning of the breath, exploration is safer and also more valuable. The former, due to the lung being at steady state and the latter due to the fact that the typical target waveform for inhalation requires a rapid pressure increase with a quick switch to stabilization, leading to a need for better understanding of dynamics in the early phases of a breath. The structure for the triangular exploration is inspired by the need for a persistent exploration strategy (similar ideas exist in \cite{dabney2020temporally}) which can capture intrinsic delay in the system. We illustrate this approach in Figure \ref{fig:exp_data}: control inputs used in our exploration policy are shown on the top, and the pressure measurements of the ventilator-lung system are shown on the bottom. Precise parameters for our exploration policy are listed in Table \ref{table:appendix-sim-data} in the Appendix.

\subsection{Model architecture}\label{sec:sim_model}

Now we describe the architectural details of our data-driven simulator. Due to the inherent differences across lungs, we opt to learn a different simulator for each of the tasks, which we can wrap into a single meta-simulator through code that selects the appropriate model based on a user's input of $R$ and $C$ parameters.

\paragraph{Training Task(s).} The simulator aims to learn the unknown dynamics of the inhalation phase. We approximate the state of the system (which is not observable to us) by the sequence of the past pressures and controls upto a history length of $H_c$ and $H_p$ respectively. The task of the simulator can now be distilled down to that of predicting the next pressure $p_{t+1}$, based on the past $H_c$ controls $u_{t},\ldots, u_{t-H_c}$ and $H_p$ pressures $p_{t},\ldots, p_{t-H_p}$. We define the training task by constructing a regression data set whose inputs come from contiguous overlapping sections of $H_p, H_c$ within the collected trajectories and the task is to predict the following pressure.

\paragraph{Boundary Model} Towards further improvement of simulator performance we found that, additional distinction needs to be provided for difference in the behavior of the dynamics during the "rise" and "stabilize" phases of an inhalation. Thus we learned a collection of individual models for the very beginning of the inhalation/episode and a general model for the rest of the inhalation, mirroring our choice of exploration policies. This proves to be very helpful as the dynamics at the very beginning of an inhalation are transient, and also extremely important to get right due to downstream effects. Concretely, our final model stitches together a list of $N_B$ boundary models and a general model, whose training tasks are as described earlier (details found in Appendix \ref{app:sim}, Table \ref{table:appendix-sim-training}). 



\section{Learning controllers from learned physics} \label{sec:control}


In this section we describe the following two controller tasks:
\begin{enumerate}
    \item {\bf Performance:} improve performance for tracking desired waveform in ISO-specified benchmarks. Specifically, we minimize the combined $L_1$ deviation from the target inhalation behavior across all target pressures on the simulator corresponding to a single lung setting of interest.
    \item {\bf Robustness:} improve performance using a {\bf single} trained controller. Specifically, we minimize the combined $L_1$ deviation from the target inhalation behavior across all target pressures \textit{and} across the simulators corresponding to \textit{several} lung settings of interest.
\end{enumerate}

\paragraph{Controller architecture.} 
Our controller is comprised of a PID baseline upon which we learn a deep network correction controlled with a regularization parameter $\lambda$. This \textit{residual} setup can be seen as a regularization against the gap between the simulator and the real dynamics. In particular this prevents the controller training from over-fitting on the simulator. We found this approach to be significantly better than the directly using the best (and perhaps over-fitted) controller on the simulator. We provide further details about the architecture and ablation studies in the Appendix. 

\subsection{Experiments}

For our experiments, we use the physical test lung to run our proposed controllers (trained on the simulators) and compare it against the PID controller that perform best on the physical lung.

To make comparisons, we compute a score for each controller on a given test lung setting (e.g., $R=5, C=50$) by averaging the $L_1$ deviation from a target pressure waveform for all inspiratory phases, and then averaging these average $L_1$ errors over six waveforms specified in \citet{ISO68844}. We choose $L_1$ as an error metric so as not to over-penalize breaths that fall short of their target pressures and to avoid engineering a new metric. We determine the best performing PID controller for a given setting by running exhaustive grid searches over $P,I,D$ coefficients for each lung setting (details for both our score and the grid searches can be found in the Appendix).



\begin{figure}[!h]
\centering 
\includegraphics[width=8cm]{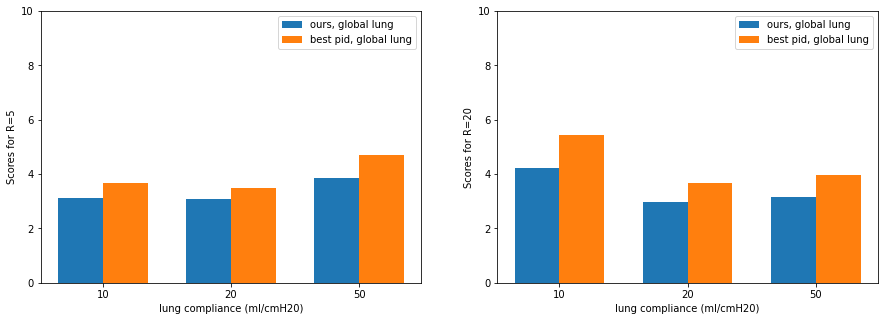}
\caption{We show that for each lung setting, the controller we trained on the simulator for that setting outperforms the best-performing PID controller found on the physical test lung.}\label{fig:results-tracking}
\end{figure}


\begin{figure}[!h]
\centering 
\includegraphics[width=8cm]{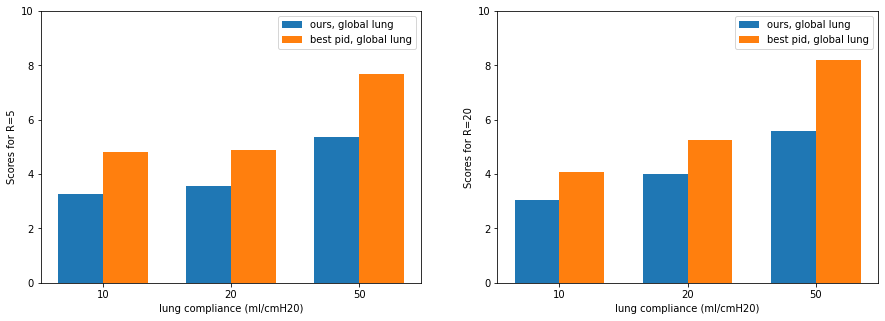}
\caption{The controller we trained on all six simulators outperforms the best PID found over the the same six settings on the physical test lung. Of note, our wins are proportionally greater when trained on all six settings whereas individual lung settings are more achievable by PID alone.}\label{fig:results-generalization}
\end{figure}

\begin{figure}[!h]
\centering 
\includegraphics[width=8cm]{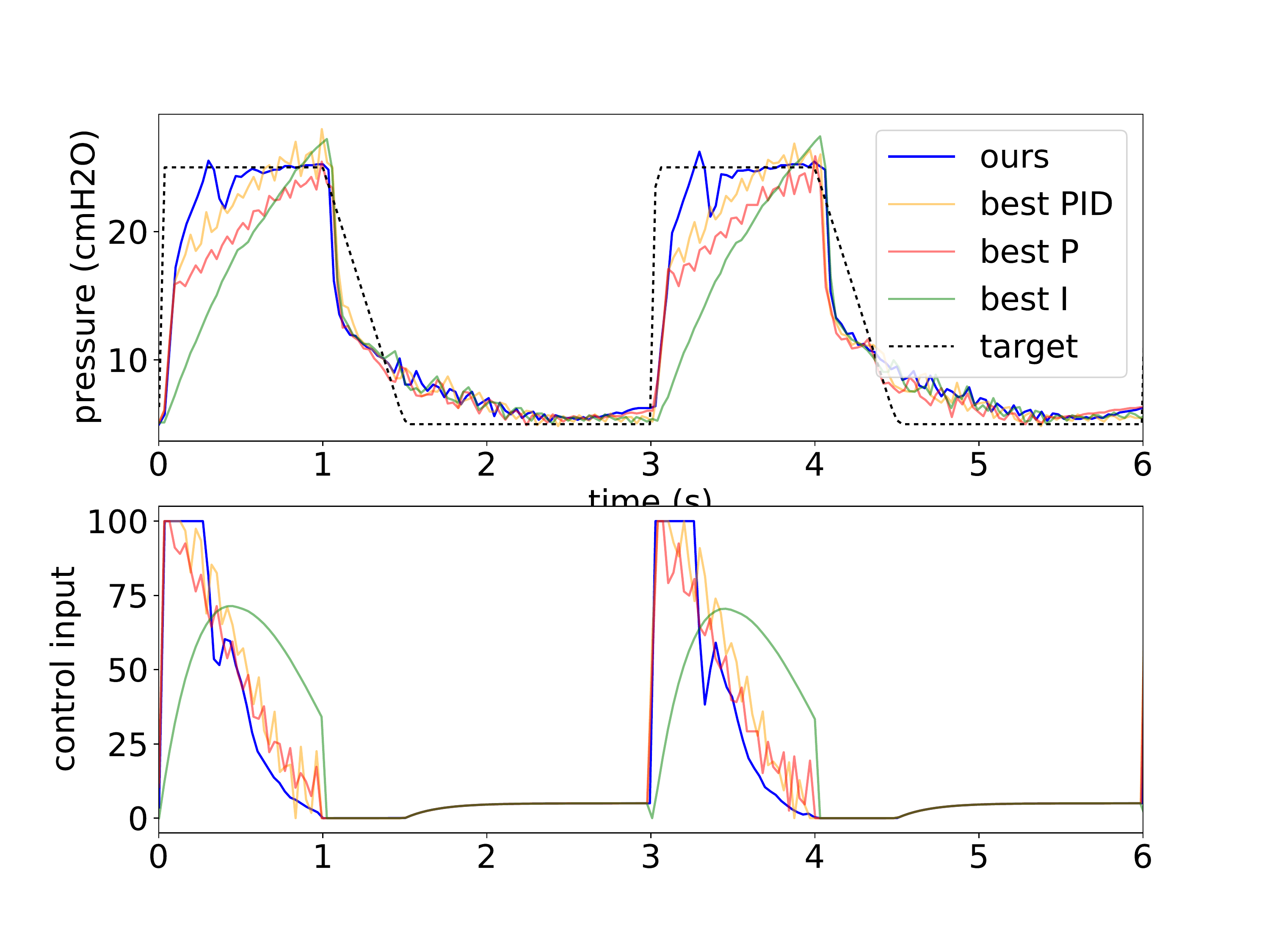}
\caption{As an example, we compare our method (learned controller on learned simulator) to the best P-only, I-only, and PID controllers relative to a target waveform (dotted line). Whereas our controller rises quickly and stays very near the target waveform, the other controllers take significantly longer to rise, overshoot, and, in the case of P-only and PID, ring the entire inspiratory phase.}\label{fig:traj}
\end{figure}


\section{Comparison of RL methods} \label{sec:benchmarks}
\begin{figure}[!h]
\centering 
\includegraphics[width=8cm]{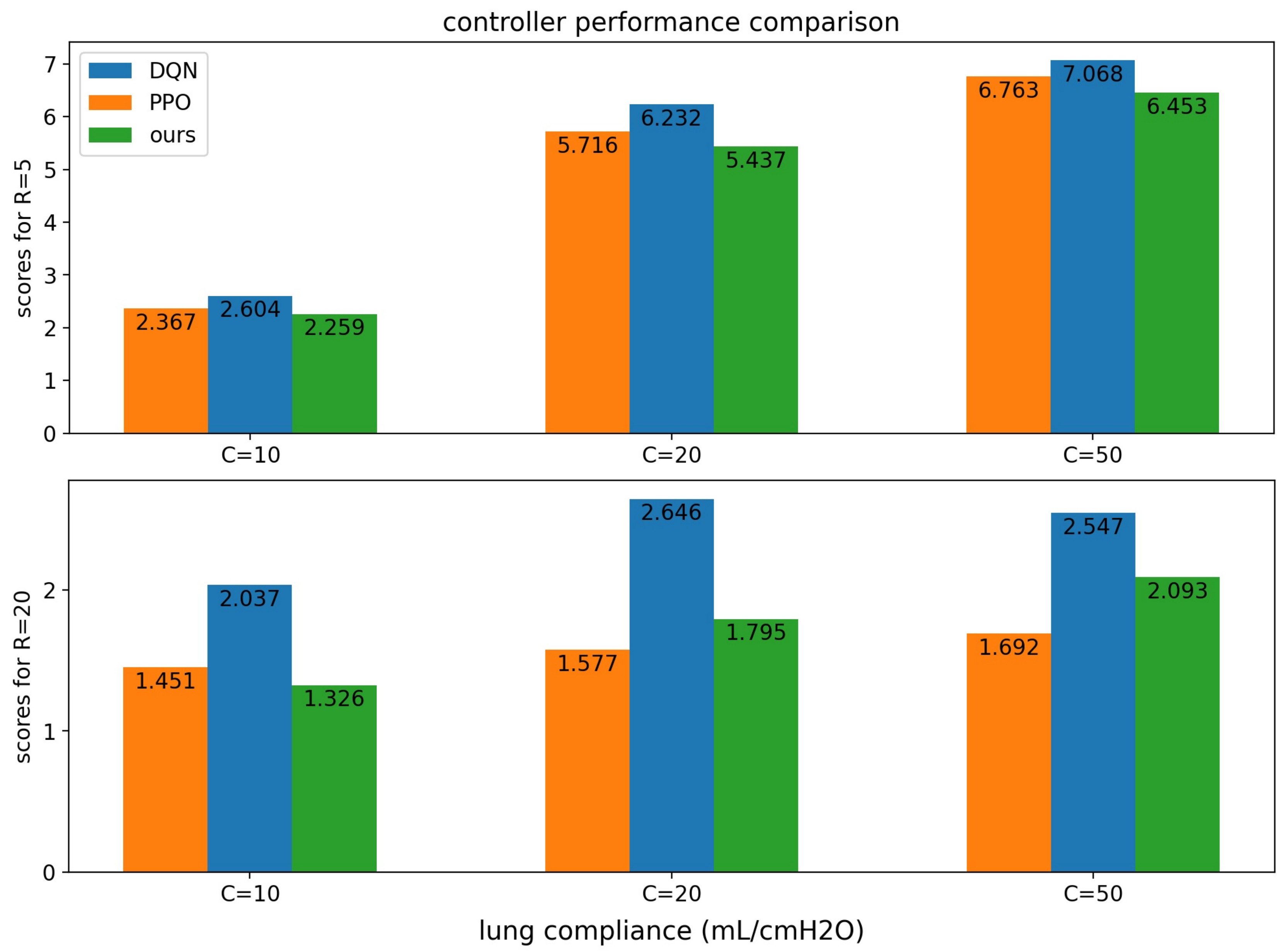}
\caption{Performance comparison of our controller with PPO/DQN. The score is calculated by average per-step L1 distance between target and achieved pressure.}\label{fig:baselines}
\end{figure}

As part of our investigation, we benchmarked and compared several Reinforcement Learning(RL) methods for policy optimization on the simulator before settling on the analytic policy gradient approach that leverages the ability to differentiate through the simulated dynamics outlined before. We consider popular RL algorithms, namely PPO~\citep{ppo} and DQN~\citep{dqn} and compare them to direct analytical policy gradient descent. These algorithms are representative of two mainstream RL paradigms, policy gradient and Q-learning, respectively. We performed experiments on simulators that represent lungs with different R,C parameters. The metric as earlier is the L1 distance between the target and achieved lung pressure per step. To ensure a fair model comparison, we used the same state featurization (as described in the previous section) for all algorithms and performed extensive hyperparameter search for our baselines during the training phase. Results are shown in Figure~\ref{fig:baselines}. Our algorithm achieves comparable scores to the baselines across all simulators.

\begin{figure}[!h]
\centering 
\includegraphics[width=8cm]{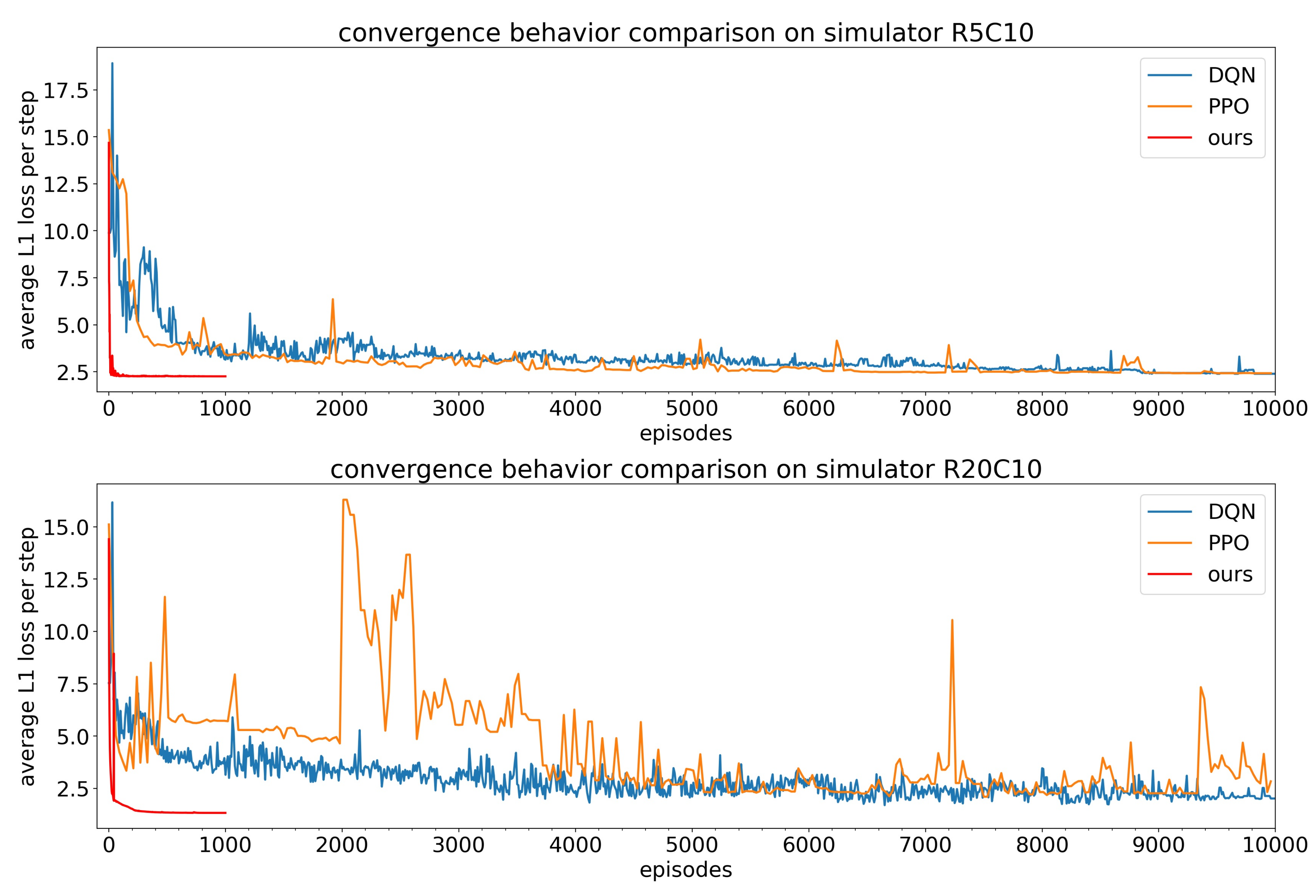}
\caption{Convergence behavior demonstration. }\label{fig:samplecomplexity}
\end{figure}

Importantly, our analytical gradient based method achieves a comparable score relative to PPO/DQN in orders of magnitude less samples. This sample-efficiency property of our algorithm can be clearly observed from  (Figure~\ref{fig:samplecomplexity}). Our method converges within ~100 episodes of training, while the other methods require tens of thousands of episodes. Further, our algorithm has a stable training process, in contrast to the notable training instability for the baselines. Furthermore, our method is robust with respect to hyperparameter tuning, unlike the baselines, which require an extensive search over hyperparameters to achieve comparable performance. This extensive hyperparameter search required by the baselines is unfeasible for use in resource-constrained or online learning scenarios, which are typical use cases for these control systems. Specifically for the results provided here, we conducted 720 trials with different hyperparameter configuration for PPO and 180 trials for DQN. In contrast, using our method only involves a few trials of standard optimizer learning rate tuning, which is the minimum effort in deep learning practices.

\section{Conclusions and future work} \label{sec:discussion}

We have presented a machine learning approach to ventilator control, demonstrating the potential of end-to-end learned controllers by obtaining improvements over industry-standard baselines.
Our main conclusions are
\begin{enumerate}
    \item The nonlinear dynamical system of lung-ventilator can be modeled by a neural network more accurately than previously studied physics-based models.
    
    \item Controllers based on deep neural networks can outperform PID controllers across multiple clinical settings (waveforms), and can generalize better across patient lungs characteristics, despite having significantly more parameters. 
    
    \item 
    Direct policy optimization for differentiable environments has potential to significantly outperform Q-learning or (standard) policy gradient methods in terms of sample and computational complexity. 
    
\end{enumerate}

There remain a number of areas to explore, mostly motivated by medical need. The lung settings we examined are by no means representative of all lung characteristics (e.g., neonatal, child, non-sedated) and lung characteristics are not static over time; a patient may improve or worsen, or begin coughing. Ventilator costs also drive further research. As an example, inexpensive valves have less consistent behavior and longer reaction times, which exacerbate bad PID behavior (e.g., overshooting, ringing), yet are crucial to bringing down costs and expanding access. Learned controllers that adapt to these deficiencies may obviate the need for such trade-offs.



\bibliography{vent}

\begin{thebibliography}{55}
\providecommand{\natexlab}[1]{#1}
\providecommand{\url}[1]{\texttt{#1}}
\expandafter\ifx\csname urlstyle\endcsname\relax
  \providecommand{\doi}[1]{doi: #1}\else
  \providecommand{\doi}{doi: \begingroup \urlstyle{rm}\Url}\fi

\bibitem[Akkaya et~al.(2019)Akkaya, Andrychowicz, Chociej, Litwin, McGrew,
  Petron, Paino, Plappert, Powell, Ribas, et~al.]{akkaya2019solving}
Ilge Akkaya, Marcin Andrychowicz, Maciek Chociej, Mateusz Litwin, Bob McGrew,
  Arthur Petron, Alex Paino, Matthias Plappert, Glenn Powell, Raphael Ribas,
  et~al.
\newblock Solving rubik's cube with a robot hand.
\newblock \emph{arXiv preprint arXiv:1910.07113}, 2019.

\bibitem[{\AA}str{\"o}m and H{\"a}gglund(1995)]{PID1}
{Karl Johan} {\AA}str{\"o}m and Tore H{\"a}gglund.
\newblock \emph{PID Controllers: Theory, Design, and Tuning}.
\newblock ISA - The Instrumentation, Systems and Automation Society, 1995.
\newblock ISBN 1-55617-516-7.

\bibitem[Bellemare et~al.(2020)Bellemare, Candido, Castro, Gong, Machado,
  Moitra, Ponda, and Wang]{bellemare2020autonomous}
Marc~G Bellemare, Salvatore Candido, Pablo~Samuel Castro, Jun Gong, Marlos~C
  Machado, Subhodeep Moitra, Sameera~S Ponda, and Ziyu Wang.
\newblock Autonomous navigation of stratospheric balloons using reinforcement
  learning.
\newblock \emph{Nature}, 588\penalty0 (7836):\penalty0 77--82, 2020.

\bibitem[{Bennett}(1993)]{pid2}
S.~{Bennett}.
\newblock Development of the pid controller.
\newblock \emph{IEEE Control Systems Magazine}, 13\penalty0 (6):\penalty0
  58--62, 1993.

\bibitem[Bertsekas(2017)]{Bertsekas17}
Dimitri~P. Bertsekas.
\newblock \emph{Dynamic Programming and Optimal Control}, volume~I.
\newblock Athena Scientific, Belmont, MA, USA, 4th edition, 2017.

\bibitem[Billings(1980)]{billings1980identification}
Stephen~A Billings.
\newblock Identification of nonlinear systems-a survey.
\newblock In \emph{IEE Proceedings D-Control Theory and Applications}, volume
  127, pages 272--285. IET, 1980.

\bibitem[Bouteloup et~al.(2020)Bouteloup, Vilsbol, Alaa, and
  Branciard]{ventlist}
Julien Bouteloup, Emmanuel Vilsbol, Asem Alaa, and Francois Branciard.
\newblock Covid-19-open-source-ventilators: List of all covid-19 open source
  ventilator initiatives.
\newblock \url{https://github.com/bneiluj/covid-19-open-source-ventilators},
  2020.

\bibitem[Chatburn(2007)]{chatburn2007classification}
Robert~L Chatburn.
\newblock Classification of ventilator modes: update and proposal for
  implementation.
\newblock \emph{Respiratory care}, 52\penalty0 (3):\penalty0 301--323, 2007.

\bibitem[Chatburn and Mireles-Cabodevila(2011)]{chatburn2011closed}
Robert~L Chatburn and Eduardo Mireles-Cabodevila.
\newblock Closed-loop control of mechanical ventilation: description and
  classification of targeting schemes.
\newblock \emph{Respiratory care}, 56\penalty0 (1):\penalty0 85--102, 2011.

\bibitem[Chebotar et~al.(2019)Chebotar, Handa, Makoviychuk, Macklin, Issac,
  Ratliff, and Fox]{chebotar2019closing}
Yevgen Chebotar, Ankur Handa, Viktor Makoviychuk, Miles Macklin, Jan Issac,
  Nathan Ratliff, and Dieter Fox.
\newblock Closing the sim-to-real loop: Adapting simulation randomization with
  real world experience.
\newblock In \emph{2019 International Conference on Robotics and Automation
  (ICRA)}, pages 8973--8979. IEEE, 2019.

\bibitem[Chen et~al.(2012)Chen, Hu, and Dai]{chen2012control}
Zheng-Long Chen, Zhao-Yan Hu, and Hou-De Dai.
\newblock Control system design for a continuous positive airway pressure
  ventilator.
\newblock \emph{Biomedical engineering online}, 11\penalty0 (1):\penalty0
  1--13, 2012.

\bibitem[Coppola et~al.(2014)Coppola, Froio, and
  Chiumello]{coppola2014protective}
Silvia Coppola, Sara Froio, and Davide Chiumello.
\newblock Protective lung ventilation during general anesthesia: is there any
  evidence?
\newblock \emph{Annual Update in Intensive Care and Emergency Medicine 2014},
  pages 159--171, 2014.

\bibitem[Cruz et~al.(2018)Cruz, Ball, Rocco, and Pelosi]{vili}
Fernanda~Ferreira Cruz, Lorenzo Ball, Patricia Rieken~Macedo Rocco, and Paolo
  Pelosi.
\newblock Ventilator-induced lung injury during controlled ventilation in
  patients with acute respiratory distress syndrome: less is probably better.
\newblock \emph{Expert Review of Respiratory Medicine}, 12\penalty0
  (5):\penalty0 403--414, 2018.
\newblock \doi{10.1080/17476348.2018.1457954}.
\newblock PMID: 29575957.

\bibitem[Dabney et~al.(2020)Dabney, Ostrovski, and
  Barreto]{dabney2020temporally}
Will Dabney, Georg Ostrovski, and Andr{\'e} Barreto.
\newblock Temporally-extended $\{$$\backslash$epsilon$\}$-greedy exploration.
\newblock \emph{arXiv preprint arXiv:2006.01782}, 2020.

\bibitem[Ferns et~al.(2005)Ferns, Panangaden, and Precup]{ferns2005metrics}
Norm Ferns, Prakash Panangaden, and Doina Precup.
\newblock Metrics for markov decision processes with infinite state spaces.
\newblock In \emph{Proceedings of the Twenty-First Conference on Uncertainty in
  Artificial Intelligence}, pages 201--208, 2005.

\bibitem[Ghassemi et~al.(2020)Ghassemi, Naumann, Schulam, Beam, Chen, and
  Ranganath]{ghassemi2020review}
Marzyeh Ghassemi, Tristan Naumann, Peter Schulam, Andrew~L Beam, Irene~Y Chen,
  and Rajesh Ranganath.
\newblock A review of challenges and opportunities in machine learning for
  health.
\newblock \emph{AMIA Summits on Translational Science Proceedings},
  2020:\penalty0 191, 2020.

\bibitem[Ha and Schmidhuber(2018)]{ha2018recurrent}
David Ha and J{\"u}rgen Schmidhuber.
\newblock Recurrent world models facilitate policy evolution.
\newblock \emph{arXiv preprint arXiv:1809.01999}, 2018.

\bibitem[Hafner et~al.(2019)Hafner, Lillicrap, Fischer, Villegas, Ha, Lee, and
  Davidson]{hafner2019learning}
Danijar Hafner, Timothy Lillicrap, Ian Fischer, Ruben Villegas, David Ha,
  Honglak Lee, and James Davidson.
\newblock Learning latent dynamics for planning from pixels.
\newblock In \emph{International Conference on Machine Learning}, pages
  2555--2565. PMLR, 2019.

\bibitem[{Hazarika} and {Swarup}(2020)]{9122946}
H.~{Hazarika} and A.~{Swarup}.
\newblock Improved performance of flow rate tracking in a ventilator using
  iterative learning control.
\newblock In \emph{2020 International Conference on Electrical and Electronics
  Engineering (ICE3)}, pages 446--451, 2020.

\bibitem[Hynes et~al.(2020)Hynes, Sapozhnikova, and Dusparic]{pidcar}
Andrew Hynes, Elena Sapozhnikova, and Ivana Dusparic.
\newblock Optimising pid control with residual policy reinforcement learning.
\newblock 12 2020.

\bibitem[IngMar(2020)]{ingmar_medical_2020}
IngMar.
\newblock Quicklung products, May 2020.
\newblock URL \url{https://www.ingmarmed.com/product/quicklung/}.

\bibitem[ISO 80601-2-80:2018()]{ISO68844}
ISO 80601-2-80:2018.
\newblock Medical electrical equipment — part 2-80: Particular requirements
  for basic safety and essential performance of ventilatory support equipment
  for ventilatory insufficiency.
\newblock Standard, International Organization for Standardization, Geneva, CH,
  July 2018.

\bibitem[Jain et~al.(2017)Jain, Kollisch-Singule, Satalin, Searles, Dombert,
  Abdel-Razek, Yepuri, Leonard, Gruessner, Andrews, Fazal, Meng, Wang, Gatto,
  Habashi, and Nieman]{vili_1}
Sumeet~V Jain, Michaela Kollisch-Singule, Joshua Satalin, Quinn Searles, Luke
  Dombert, Osama Abdel-Razek, Natesh Yepuri, Antony Leonard, Angelika
  Gruessner, Penny Andrews, Fabeha Fazal, Qinghe Meng, Guirong Wang, Louis~A
  Gatto, Nader~M Habashi, and Gary~F Nieman.
\newblock The role of high airway pressure and dynamic strain on
  ventilator-induced lung injury in a heterogeneous acute lung injury model.
\newblock \emph{Intensive Care Med Exp.5(1):25.}, 2017.

\bibitem[Kacmarek(2011)]{kacmarek11}
Robert Kacmarek.
\newblock The mechanical ventilator: Past, present, and future.
\newblock \emph{Respiratory care}, 56:\penalty0 1170--80, 08 2011.
\newblock \doi{10.4187/respcare.01420}.

\bibitem[Kaiser et~al.(2019)Kaiser, Babaeizadeh, Milos, Osinski, Campbell,
  Czechowski, Erhan, Finn, Kozakowski, Levine, et~al.]{kaiser2019model}
Lukasz Kaiser, Mohammad Babaeizadeh, Piotr Milos, Blazej Osinski, Roy~H
  Campbell, Konrad Czechowski, Dumitru Erhan, Chelsea Finn, Piotr Kozakowski,
  Sergey Levine, et~al.
\newblock Model-based reinforcement learning for {A}tari.
\newblock \emph{arXiv preprint arXiv:1903.00374}, 2019.

\bibitem[LaChance et~al.(2020)LaChance, Zajdel, Schottdorf, Saunders, Dvali,
  Marshall, Seirup, Notterman, and Cohen]{pvp2020}
J.~LaChance, Tom~J. Zajdel, Manuel Schottdorf, Jonny~L. Saunders, Sophie Dvali,
  Chase Marshall, Lorenzo Seirup, Daniel~A. Notterman, and Daniel~J. Cohen.
\newblock Pvp1 -- the people’s ventilator project: A fully open, low-cost,
  pressure-controlled ventilator, 2020.

\bibitem[Marini and Crooke(1993)]{PMID:8420408}
JJ~Marini and PS~Crooke.
\newblock A general mathematical model for respiratory dynamics relevant to the
  clinical setting.
\newblock \emph{The American review of respiratory disease}, 147\penalty0
  (1):\penalty0 14—24, January 1993.
\newblock ISSN 0003-0805.
\newblock \doi{10.1164/ajrccm/147.1.14}.
\newblock URL \url{https://doi.org/10.1164/ajrccm/147.1.14}.

\bibitem[McIntyre(1986)]{canadian1986canadian}
John~WR McIntyre.
\newblock Anaesthesia breathing circuits.
\newblock \emph{The Canadian Anaesthetists' Society Journal: Journal de la
  Soci{\'e}t{\'e} Canadienne Des Anesth{\'e}sistes}, 33\penalty0 (2), 1986.

\bibitem[Mellott et~al.()Mellott, Grap, Munro, Sessler, and Wetzel]{asynchrony}
Karen~G. Mellott, Mary~Jo Grap, Cindy~L. Munro, Curtis~N. Sessler, and Paul~A.
  Wetzel.
\newblock Patient-ventilator dyssynchrony: clinical significance and
  implications for practice.
\newblock \emph{Crit Care Nurse. 2009;29(6):41-55.}

\bibitem[Meng et~al.(2020)Meng, Qiu, Wan, Ai, Xue, Guo, Deshpande, Zhang, Meng,
  Tong, et~al.]{meng2020intubation}
Lingzhong Meng, Haibo Qiu, Li~Wan, Yuhang Ai, Zhanggang Xue, Qulian Guo, Ranjit
  Deshpande, Lina Zhang, Jie Meng, Chuanyao Tong, et~al.
\newblock Intubation and ventilation amid the covid-19 outbreak: Wuhan’s
  experience.
\newblock \emph{Anesthesiology}, 132\penalty0 (6):\penalty0 1317--1332, 2020.

\bibitem[Mnih et~al.(2013)Mnih, Kavukcuoglu, Silver, Graves, Antonoglou,
  Wierstra, and Riedmiller]{dqn}
Volodymyr Mnih, Koray Kavukcuoglu, David Silver, Alex Graves, Ioannis
  Antonoglou, Daan Wierstra, and Martin Riedmiller.
\newblock Playing atari with deep reinforcement learning.
\newblock \emph{arXiv preprint arXiv:1312.5602}, 2013.

\bibitem[M{\"o}hlenkamp and Thiele(2020)]{mohlenkamp2020ventilation}
Stefan M{\"o}hlenkamp and Holger Thiele.
\newblock Ventilation of covid-19 patients in intensive care units.
\newblock \emph{Herz}, 45\penalty0 (4):\penalty0 329--331, 2020.

\bibitem[Nadeem(2021)]{nadeem_2021}
Nimra Nadeem.
\newblock Learning to breathe: Physics v. ml-based lung simulations for control
  of medical ventilators, 2021.

\bibitem[Oto et~al.(2021)Oto, Annesi, and Foley]{patientventasynch}
Brandon Oto, Janet Annesi, and Raymond~J Foley.
\newblock Patient--ventilator dyssynchrony in the intensive care unit: A
  practical approach to diagnosis and management.
\newblock \emph{Anaesthesia and Intensive Care}, 49\penalty0 (2):\penalty0
  86--97, 2021.
\newblock \doi{10.1177/0310057X20978981}.
\newblock URL \url{https://doi.org/10.1177/0310057X20978981}.
\newblock PMID: 33906464.

\bibitem[Parmley et~al.(1972)Parmley, Tahir, Dascomb, and
  Adriani]{parmley1972disposable}
John~B Parmley, Ashiq~H Tahir, Harry~E Dascomb, and John Adriani.
\newblock Disposable versus reusable rebreathing circuits: Advantages,
  disadvantages, hazards and bacteiiologic studies.
\newblock \emph{Anesthesia \& Analgesia}, 51\penalty0 (6):\penalty0 888--894,
  1972.

\bibitem[Prasad et~al.(2017)Prasad, Cheng, Chivers, Draugelis, and
  Engelhardt]{prasad2017reinforcement}
Niranjani Prasad, Li-Fang Cheng, Corey Chivers, Michael Draugelis, and
  Barbara~E Engelhardt.
\newblock A reinforcement learning approach to weaning of mechanical
  ventilation in intensive care units, 2017.

\bibitem[Punjani and Abbeel(2015)]{helicopter}
Ali Punjani and Pieter Abbeel.
\newblock Deep learning helicopter dynamics models.
\newblock In \emph{2015 IEEE International Conference on Robotics and
  Automation (ICRA)}, pages 3223--3230, 2015.
\newblock \doi{10.1109/ICRA.2015.7139643}.

\bibitem[Rajeswaran et~al.(2016)Rajeswaran, Ghotra, Ravindran, and
  Levine]{rajeswaran2016epopt}
Aravind Rajeswaran, Sarvjeet Ghotra, Balaraman Ravindran, and Sergey Levine.
\newblock Epopt: Learning robust neural network policies using model ensembles.
\newblock \emph{arXiv preprint arXiv:1610.01283}, 2016.

\bibitem[Rees et~al.(2006)Rees, Aller{\o}d, Murley, Zhao, Smith, Kj{\ae}rgaard,
  Thorgaard, and Andreassen]{rees2006using}
Stephen~Edward Rees, Charlotte Aller{\o}d, David Murley, Yichun Zhao,
  Bram~Wallace Smith, S~Kj{\ae}rgaard, P~Thorgaard, and Steen Andreassen.
\newblock Using physiological models and decision theory for selecting
  appropriate ventilator settings.
\newblock \emph{Journal of clinical monitoring and computing}, 20\penalty0
  (6):\penalty0 421--429, 2006.

\bibitem[Rittayamai et~al.(2015)Rittayamai, Katsios, Beloncle, Friedrich,
  Mancebo, and Brochard]{rittayamai2015pressure}
Nuttapol Rittayamai, Christina~M Katsios, Fran{\c{c}}ois Beloncle, Jan~O
  Friedrich, Jordi Mancebo, and Laurent Brochard.
\newblock Pressure-controlled vs volume-controlled ventilation in acute
  respiratory failure: a physiology-based narrative and systematic review.
\newblock \emph{Chest}, 148\penalty0 (2):\penalty0 340--355, 2015.

\bibitem[Schoukens and Ljung(2019)]{schoukens2019nonlinear}
Johan Schoukens and Lennart Ljung.
\newblock Nonlinear system identification: A user-oriented roadmap, 2019.

\bibitem[Schrittwieser et~al.(2020)Schrittwieser, Antonoglou, Hubert, Simonyan,
  Sifre, Schmitt, Guez, Lockhart, Hassabis, Graepel,
  et~al.]{schrittwieser2020mastering}
Julian Schrittwieser, Ioannis Antonoglou, Thomas Hubert, Karen Simonyan,
  Laurent Sifre, Simon Schmitt, Arthur Guez, Edward Lockhart, Demis Hassabis,
  Thore Graepel, et~al.
\newblock Mastering atari, go, chess and shogi by planning with a learned
  model.
\newblock \emph{Nature}, 588\penalty0 (7839):\penalty0 604--609, 2020.

\bibitem[Schulman et~al.(2017)Schulman, Wolski, Dhariwal, Radford, and
  Klimov]{ppo}
John Schulman, Filip Wolski, Prafulla Dhariwal, Alec Radford, and Oleg Klimov.
\newblock Proximal policy optimization algorithms.
\newblock \emph{arXiv preprint arXiv:1707.06347}, 2017.

\bibitem[Shi et~al.(2020)Shi, Wang, Xie, and Su]{PPR:PPR169448}
Peng Shi, Na~Wang, Fei Xie, and Hang Su.
\newblock Self-adjusting ventilator control strategy based on pid, 2020.
\newblock URL \url{https://doi.org/10.21203/rs.3.rs-31632/v1}.

\bibitem[Silver et~al.(2019)Silver, Allen, Tenenbaum, and Kaelbling]{rpl}
Tom Silver, Kelsey Allen, Josh Tenenbaum, and Leslie Kaelbling.
\newblock Residual policy learning, 2019.

\bibitem[Taylor and Stone(2009)]{taylor2009transfer}
Matthew~E Taylor and Peter Stone.
\newblock Transfer learning for reinforcement learning domains: A survey.
\newblock \emph{Journal of Machine Learning Research}, 10\penalty0 (7), 2009.

\bibitem[Tedrake(2020)]{tedrake}
Russ Tedrake.
\newblock \emph{Underactuated Robotics: Algorithms for Walking, Running,
  Swimming, Flying, and Manipulation (Course Notes for MIT 6.832)}.
\newblock 2020.

\bibitem[van Kaam et~al.(2019)van Kaam, De~Luca, Hentschel, Hutten, Sindelar,
  Thome, and Zimmermann]{van2019modes}
Anton~H van Kaam, Dani{\`e}la De~Luca, Roland Hentschel, Jeroen Hutten, Richard
  Sindelar, Ulrich Thome, and Luc~JI Zimmermann.
\newblock Modes and strategies for providing conventional mechanical
  ventilation in neonates.
\newblock \emph{Pediatric research}, pages 1--6, 2019.

\bibitem[Vishwanathan et~al.(2007)Vishwanathan, Smola, and
  Vidal]{vishwanathan2007binet}
SVN Vishwanathan, Alexander~J Smola, and Ren{\'e} Vidal.
\newblock Binet-cauchy kernels on dynamical systems and its application to the
  analysis of dynamic scenes.
\newblock \emph{International Journal of Computer Vision}, 73\penalty0
  (1):\penalty0 95--119, 2007.

\bibitem[Wunsch(2020)]{wunsch2020mechanical}
Hannah Wunsch.
\newblock Mechanical ventilation in covid-19: interpreting the current
  epidemiology, 2020.

\bibitem[Yu et~al.(2019)Yu, Liu, and Zhao]{yu2019inverse}
Chao Yu, Jiming Liu, and Hongyi Zhao.
\newblock Inverse reinforcement learning for intelligent mechanical ventilation
  and sedative dosing in intensive care units.
\newblock \emph{BMC medical informatics and decision making}, 19\penalty0
  (2):\penalty0 111--120, 2019.

\bibitem[Yu et~al.(2020{\natexlab{a}})Yu, Liu, and Nemati]{yu2020reinforcement}
Chao Yu, Jiming Liu, and Shamim Nemati.
\newblock Reinforcement learning in healthcare: A survey, 2020{\natexlab{a}}.

\bibitem[Yu et~al.(2020{\natexlab{b}})Yu, Ren, and Dong]{yu2020supervised}
Chao Yu, Guoqi Ren, and Yinzhao Dong.
\newblock Supervised-actor-critic reinforcement learning for intelligent
  mechanical ventilation and sedative dosing in intensive care units.
\newblock \emph{BMC medical informatics and decision making}, 20\penalty0
  (3):\penalty0 1--8, 2020{\natexlab{b}}.

\bibitem[Zhou et~al.(1996)Zhou, Doyle, and Glover]{kemin}
Kemin Zhou, John~C. Doyle, and Keith Glover.
\newblock \emph{Robust and Optimal Control}.
\newblock Prentice-Hall, Inc., USA, 1996.
\newblock ISBN 0134565673.

\bibitem[Ziegler et~al.(1942)Ziegler, Nichols, et~al.]{ziegler1942optimum}
John~G Ziegler, Nathaniel~B Nichols, et~al.
\newblock Optimum settings for automatic controllers.
\newblock \emph{trans. ASME}, 64\penalty0 (11), 1942.

\end{thebibliography}

\clearpage
\onecolumn
\appendix

\section{Data collection}
\label{app:data}
\subsection{PID residual exploration} The following table describes the settings for determining policies $(a)$ and $(b)$ for collecting simulator training data as described in Section \ref{sec:sim}.
\label{app:data-sim-explore}
\hfill \break
\begin{table}[!h]
    \centering
    \begin{tabular}{|c|c|c|c|c|c|c|}
        \hline
        $(R, C)$ & $(P, I, D)$ & $(c^a_{\min}, c^a_{\max})$ & $(t^a_{\min}, t^a_{\max})$ & $(c^b_{\min}, c^b_{\max})$ & $(t^b_{\min}, t^b_{\max})$ & $p_a$\\
        \hline
        (5, 10) & (1, 0.5, 0) &  (50, 100) & (0.3, 0.6) & (-20, 40) & (0.1, 0.5) & 0.25 \\
        (5, 20) & (1, 3, 0) &  (50, 100) & (0.4, 0.8) & (-20, 60) & (0.1, 0.5) & 0.25 \\
        (5, 50) & (2, 4, 0) &  (75, 100) & (1.0, 1.5) & (-20, 60) & (0.1, 0.5) & 0.25 \\
        (20, 10) & (1, 0.5, 0) &  (50, 100) & (0.3, 0.6) & (-20, 40) & (0.1, 0.5) & 0.25 \\
        (20, 20) & (0, 3, 0) &  (30, 60) & (0.5, 1.0) & (-20, 40) & (0.1, 0.5) & 0.25 \\
        (20, 50) & (0, 4, 0) &  (70, 100) & (1.0, 1.5) & (-20, 40) & (0.1, 0.5) & 0.25 \\
        \hline
    \end{tabular}
    \caption{Parameters for exploring in the boundary of a PID controller}
    \label{table:appendix-sim-data}
\end{table}

\subsection{PID grid search}
\label{app:data-best-pid}

For each lung setting, we run a grid of settings over $P$, $I$, and $D$ (with values $[0.0, 0.1, 0.2, 0.3, 0.4, 0.5, 0.6, 0.7, 0.8, 0.9, $ $1.0, 2.0, 3.0, 4.0, 5.0, 6.0, 7.0, 8.0, 9.0, 10.0]$ each). For each grid point, we target six different waveforms (with identical PEEP and breaths per minute, but varying PIP over $[10, 15, 20, 25, 30, 35]$ cmH2O. This gives us 2,400 trajectories for each lung setting. We determine a score for the run by averaging the L1 loss between the actual and target pressures, ignoring the first breath. Each run lasts 300 time steps (approximately 9 seconds, or three breaths), which we have found to give sufficiently consistent results compared with a longer run.

Of note, some of our coefficients reach our maximum grid setting (i.e., 10.0). We explored going beyond 10 but found that performance actually degrades quickly since a quickly rising pressure is offset by subsequent overshooting and/or ringing.
\hfill \break

\begin{table}[H]
\centering
\begin{tabular}{ |c|c|c|c| } 
 \hline
  $(R, C)$ & $P$ & $I$ & $D$\\
  \hline
  (5, 10) & 10.0 & 0.2 & 0.0\\
(5, 20) & 10.0 & 10.0 & 0.0\\
(5, 50) & 10.0 & 10.0 & 0.0\\
(20, 10) & 8.0 & 1.0 & 0.0\\
(20, 20) & 5.0 & 10.0 & 0.0\\
(20, 50) & 5.0 & 10.0 & 0.0\\
\hline
\end{tabular}
\caption{$P$ and $I$ coefficients that give the best L1 controller performance relative to the target waveform averaged across the six waveforms associated with $PIP=[10, 15, 20, 25, 30, 35]$.}
\label{table:appendix-best-pid}
\end{table}

\section{Simulator details}
\label{app:sim}

\subsection{Evaluation}
\label{app:sim-evaluation}
\paragraph{Open-loop test.} To validate the simulator's performance, we hold out 20\% of the trajectory data we collected including residual exploration. We run the exact sequence of controls derived from the lung execution on the simulator. We define the point-wise error to be the absolute value of the distance between the pressure observed on the real lung and the corresponding output of the simulator, i.e. $\text{err}_t = \lvert p_t^{sim} - p_t^{lung} \rvert$. We assess the MAE loss corresponding to the errors accumulated across all test trajectories.

The following table contains the optimal objective values achieved via the above training and evaluation along with an architecture search over the parameters $H_p$ (pressure window), $H_c$ (control window), $W$ (width) , $d$ (depth), $N_B$ (number of boundary models), 

\begin{table}[]
    \centering
    \begin{tabular}{|c|c|c|c|c|c|c|}
\toprule
(R,C) &  Open-loop Average MAE &  $d$ &  $N_B$ &  $H_p$ &  $W$ &  $H_c$ \\
\midrule
(5,10) &         0.64 &               9.0 &                  1.0 &       5.0 &              150.0 &      10.0 \\
(5,20) &         0.72 &               6.0 &                  1.0 &       5.0 &              100.0 &       5.0 \\
(5,50) &         0.39 &               6.0 &                  1.0 &      10.0 &              150.0 &      10.0 \\
(20, 10) &         0.72 &               9.0 &                  1.0 &      10.0 &              100.0 &      10.0 \\
(20, 20) &         0.60 &               9.0 &                  1.0 &      10.0 &              150.0 &      10.0 \\
(20, 50) &         0.85 &               9.0 &                  1.0 &      10.0 &              150.0 &      10.0 \\
\bottomrule
\end{tabular}
\caption{Mean absolute error for the open-loop test under each lung setting under the optimal architectural parameters.}
    \label{table:appendix-sim-training}
\end{table}

\begin{table}[]
\centering
\begin{tabular}{ccccc}
    \begin{tabular}{|c|c|}
\toprule
 $d$ &  MAE \\
\midrule
              3.0 &         0.696342 \\
              6.0 &         0.649357 \\
              9.0 &         0.643421 \\
\bottomrule
\end{tabular} &

\begin{tabular}{|c|c|}
\toprule
 $N_B$ &  MAE \\
\midrule
                 1.0 &         0.643421 \\
                 3.0 &         0.676724 \\
                 5.0 &         0.718358 \\
\bottomrule
\end{tabular} &

\begin{tabular}{|c|c|}
\toprule
 $H_p$ &  MAE \\
\midrule
      3.0 &         0.649357 \\
      5.0 &         0.643421 \\
     10.0 &         0.647772 \\
\bottomrule
\end{tabular} &

\begin{tabular}{|c|c|}
\toprule
 $W$ &  MAE \\
\midrule
              50.0 &         0.679061 \\
             100.0 &         0.650625 \\
             150.0 &         0.643421 \\
\bottomrule
\end{tabular} &

\begin{tabular}{|c|c|}
\toprule
 $H_c$ &  MAE \\
\midrule
      3.0 &         0.675960 \\
      5.0 &         0.647772 \\
     10.0 &         0.643421 \\
\bottomrule
\end{tabular}
\end{tabular}
    \caption{Open-Loop Errors across multiple dimensions of the architecture search for one lung setting R5C10. We see while more expressive networks and featurizations lead to gains, the relative gains plateau quickly. Similar trends are observed across lung settings.}
    \label{tab:my_label}
\end{table}

\paragraph{Trajectory comparison.} In addition to the open-loop test, we compare the true trajectories to simulated ones as described in Section \ref{sec:sim}.
\hfill \break
\begin{table}[H]
    \centering
    \begin{tabular}{ccc}
        \includegraphics[width=55mm]{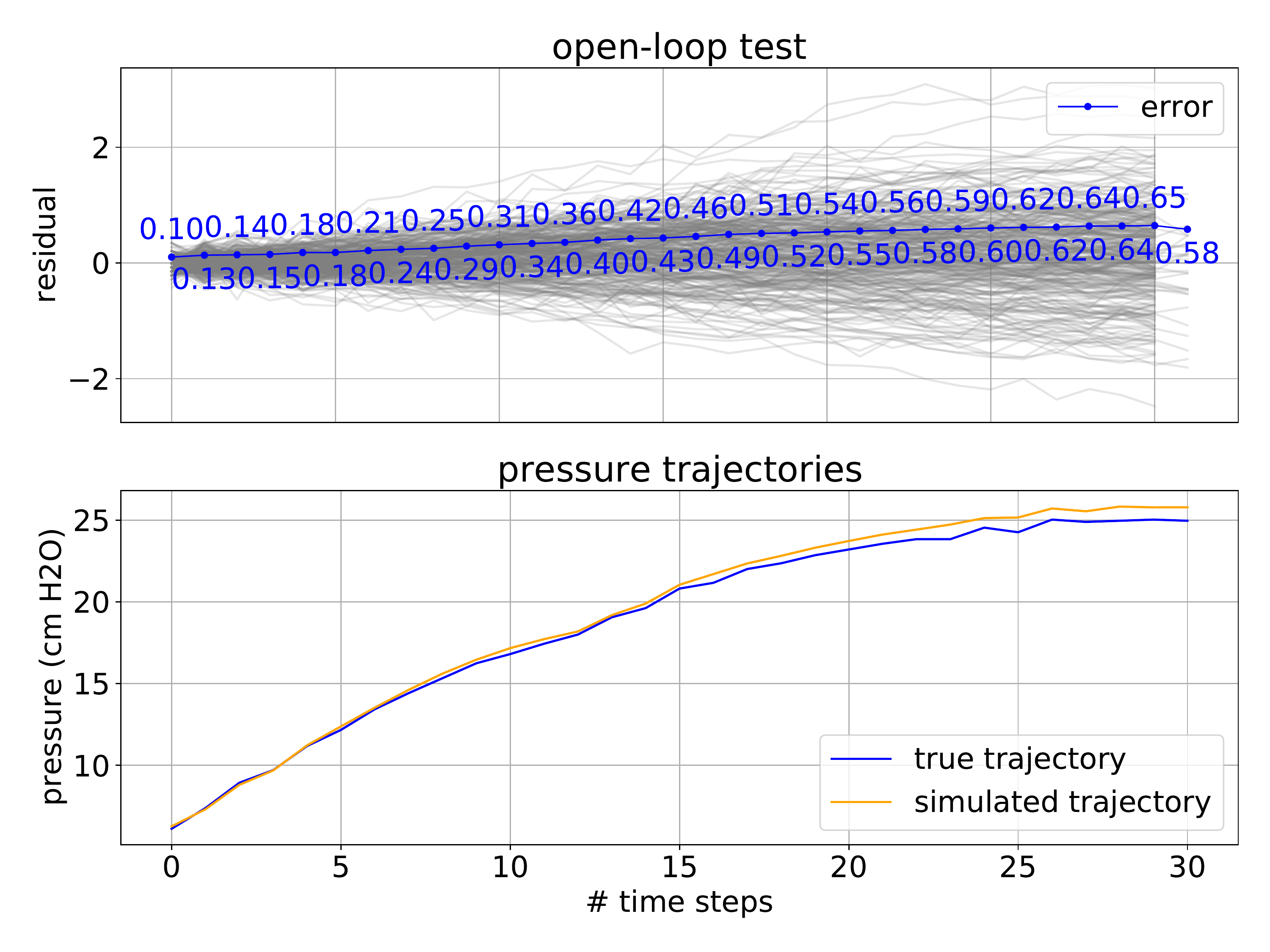} &
        \includegraphics[width=55mm]{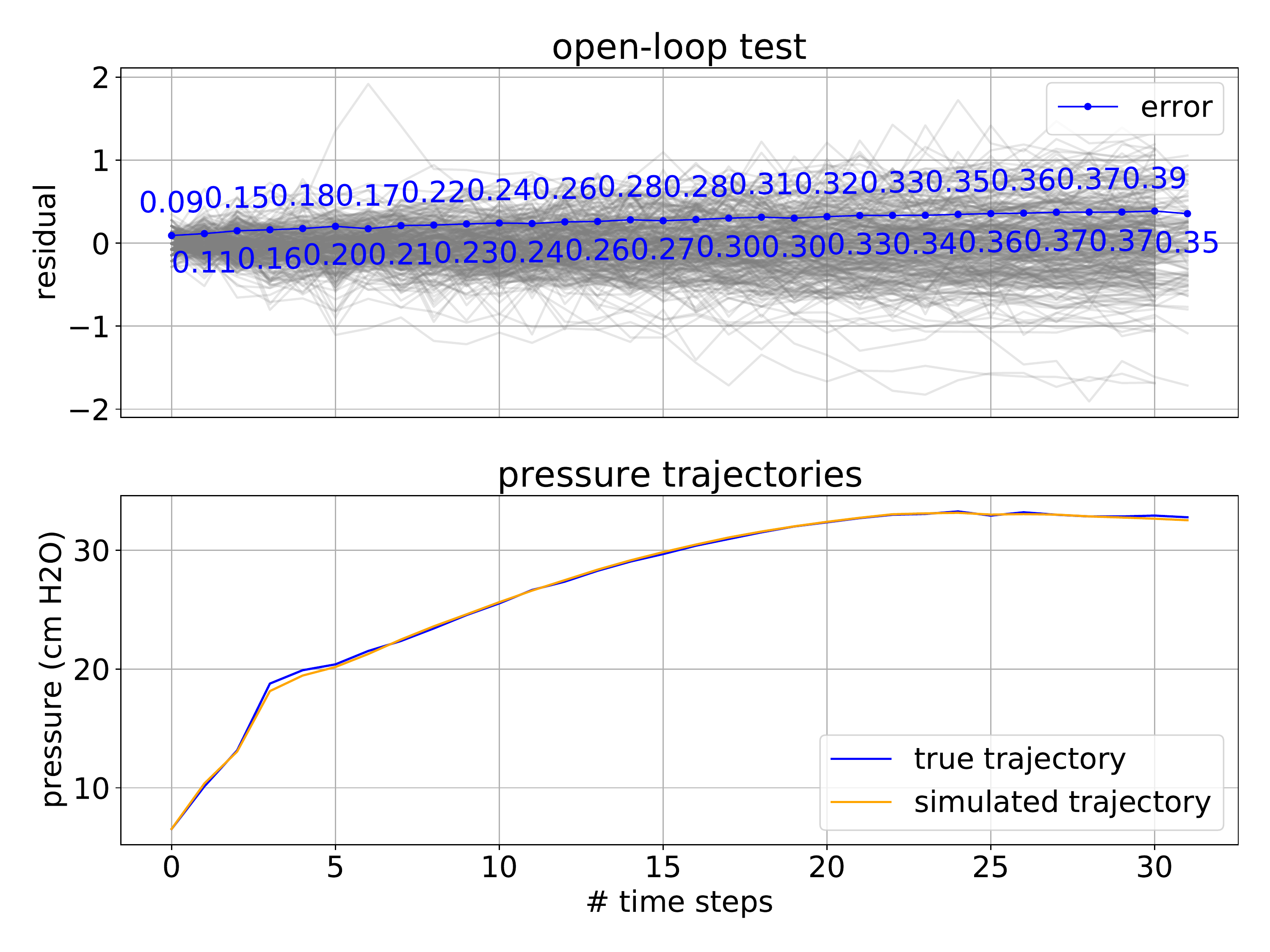} &
        \includegraphics[width=55mm]{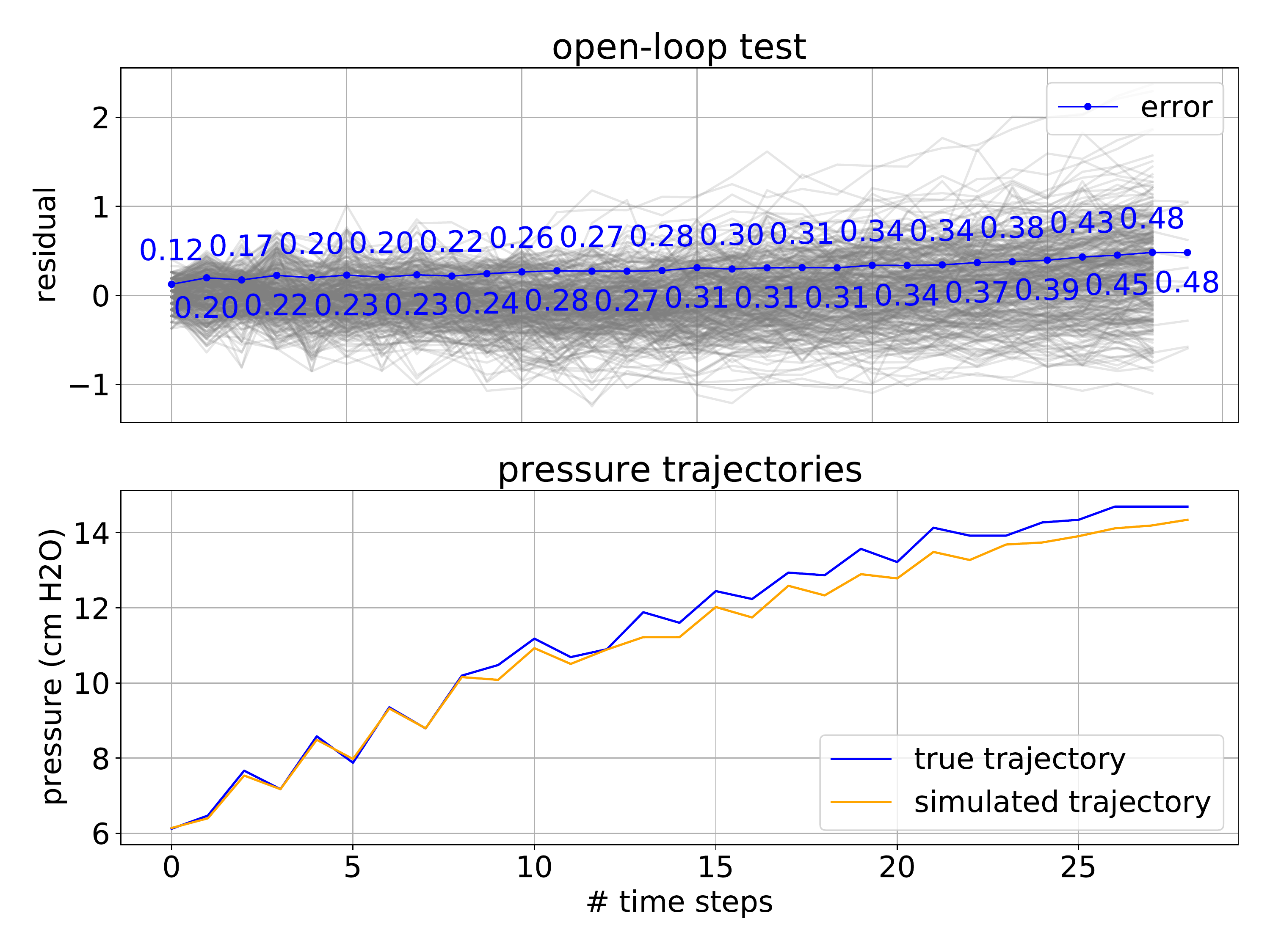}  \\
        \small R=5, C=10 & R=5, C=20 & R=5, C=50 \\
        \includegraphics[width=55mm]{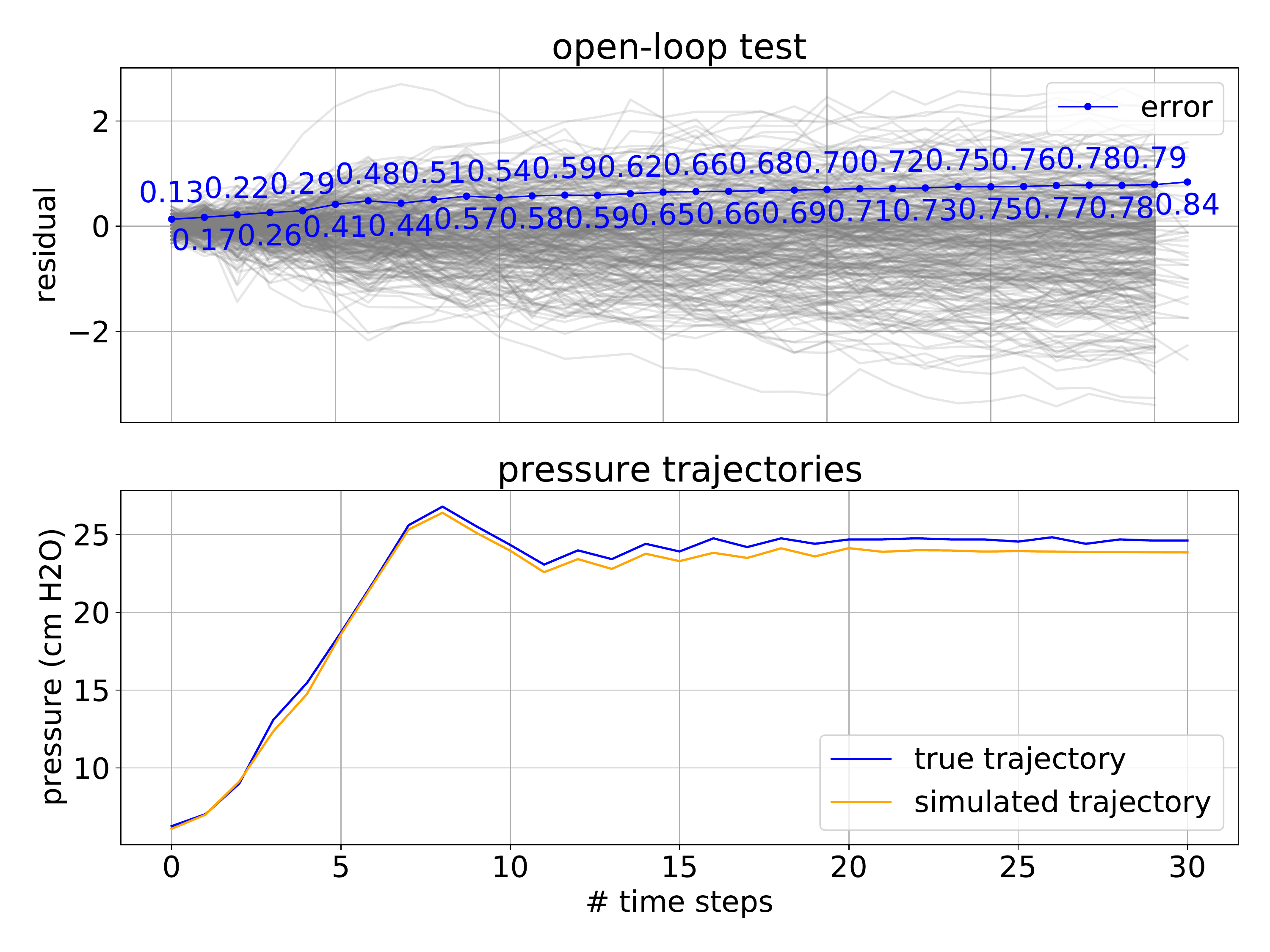} &
        \includegraphics[width=55mm]{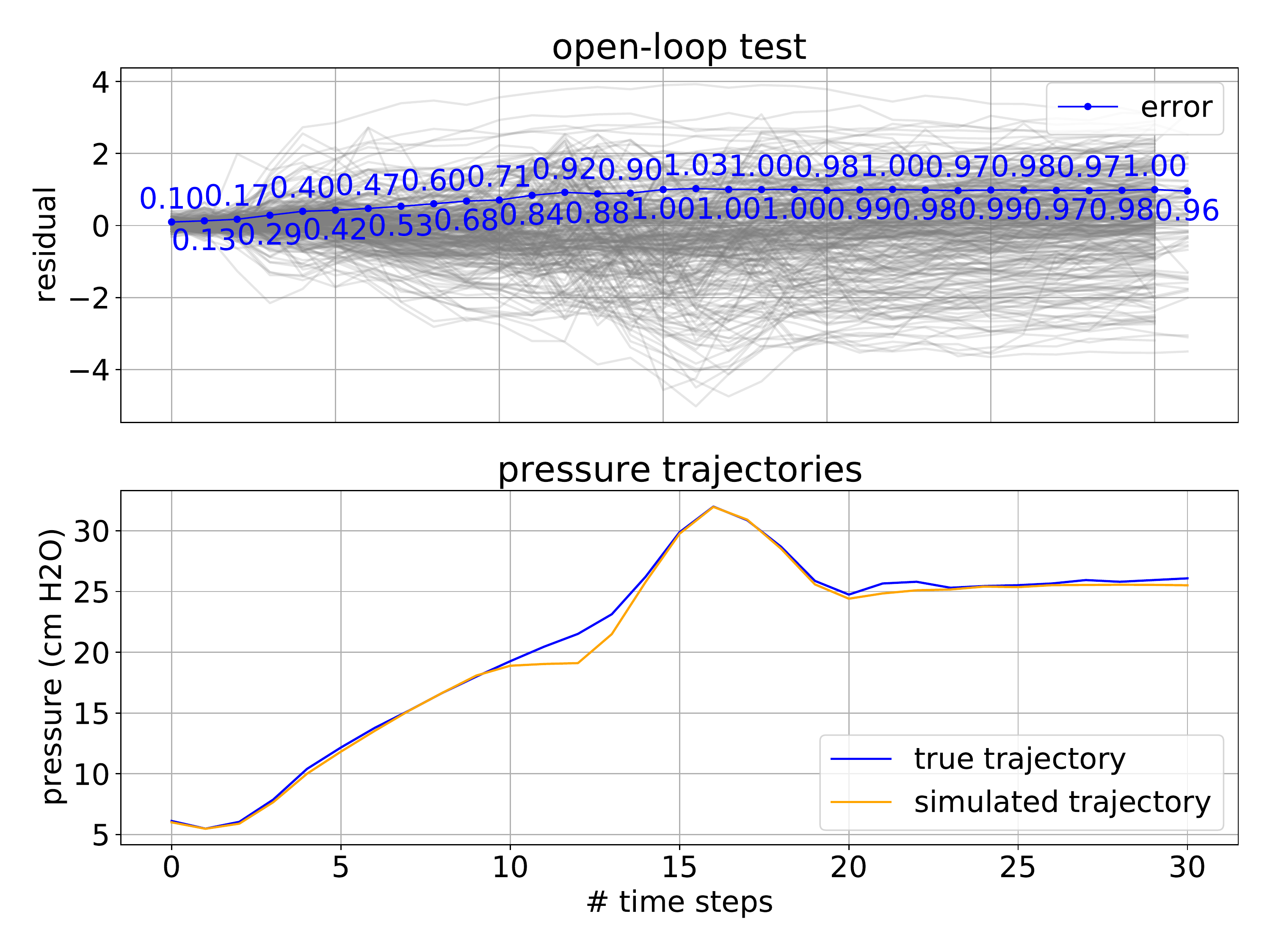} &
        \includegraphics[width=55mm]{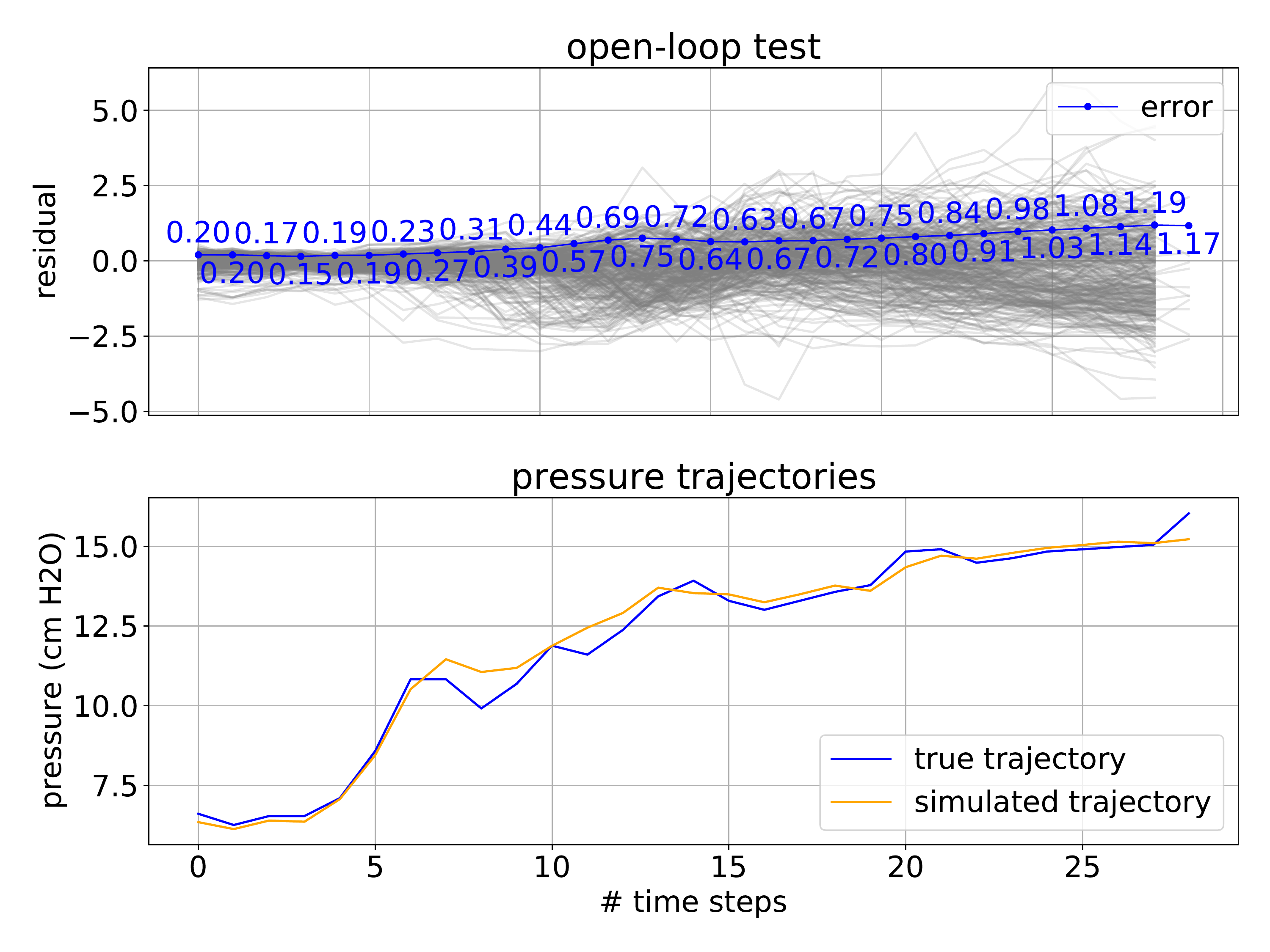}  \\
        \small R=20, C=10 & R=20, C=20 & R=20, C=50
    \end{tabular}
    \caption{We plot both open-loop testing and pressure trajectories for each of the six simulators corresponding to the six lung settings under consideration. These plots are described more in Section \ref{sec:sim}}
    \label{table:appendix-sim-plots}
\end{table}


\section{Controller details}
\label{app:controller}
\subsection{Training hyperparameters}
\label{app:controller-hyperparmeters}
We use an initial learning rate of $10^{-1}$ and weight decay $10^{-5}$ over 30 epochs.

\subsection{Training a controller across multiple simulators}
\label{app:controller-global}
To the generalization task, we train controllers across multiple simulators corresponding to the lung settings ($R = 20$, $C = [10, 20, 50]$ in our case). For each target waveform (there are six, one for each PIP in $[10, 15, 20, 25, 30, 35]$ cmH2O) and each simulator, we train the controller round-robin (i.e., one after another sequentially) once per epoch. We zero out the gradients between each epoch.

\end{document}